%% file: acl_latex.tex
\documentclass[11pt]{article}

\usepackage[preprint]{acl}

\usepackage{times}
\usepackage{latexsym}

\usepackage[T1]{fontenc}

\usepackage[utf8]{inputenc}

\usepackage{microtype}

\usepackage{inconsolata}

\usepackage{graphicx}

\usepackage[utf8]{inputenc} 
\usepackage[T1]{fontenc}    
\usepackage{hyperref}       
\usepackage{url}            
\usepackage{booktabs}       
\usepackage{amsfonts}       
\usepackage{nicefrac}       
\usepackage{microtype}      
\usepackage{xcolor}         
\definecolor{curveblue}{RGB}{44,114,181}
\definecolor{curvered}{RGB}{196,43,42}

\usepackage{graphicx}
\usepackage{amsmath}
\usepackage{wrapfig}
\usepackage{tabularx}
\usepackage{amsthm}
\usepackage{amssymb}
\usepackage{booktabs}
\usepackage{algorithm}
\usepackage{algorithmic}
\usepackage{booktabs}
\usepackage{pifont}
\usepackage{colortbl}
\usepackage[most]{tcolorbox} 
\usepackage{enumitem}
\usepackage{longtable}
\usepackage{multirow}
\definecolor{rowblue}{HTML}{E6F3FF} 
\definecolor{textred}{HTML}{FF3333} 
\definecolor{textgreen}{HTML}{009900} 

\title{
SAAS: Self-Aware Reinforcement Learning for Over-Search Mitigation in Agentic Search}

\author{
  Yunbo Tang$^{1,\dagger}$, Chengyi Yang$^{1,\dagger}$, Shiyu Liu$^{1}$, Zhishang Xiang$^{1}$, Zerui Chen$^{1}$, \\
  \textbf{Qinggang Zhang}$^{2,}$\thanks{Corresponding authors.}, \textbf{Jinsong Su}$^{1,}$\footnotemark[1] \\
  $^{1}$School of Informatics, Xiamen University \\
  $^{2}$School of Artificial Intelligence, Jilin University \\
  \texttt{tangyunbo@stu.xmu.edu.cn; yangchengyi@stu.xmu.edu.cn;} \\
  \texttt{qinggangzhang@jlu.edu.cn; jssu@xmu.edu.cn}
}

\begin{document}
\maketitle

\insert\footins{\footnotesize$^\dagger$Equal contribution.}
\begin{abstract}

Agentic search enables LLMs to solve complex multi-hop questions through iterative reasoning and external search. Despite the effectiveness, these systems often suffer from a critical limitation in practice: agents fail to recognize their own knowledge boundaries, blindly triggering searches when internal knowledge suffices and failing to terminate search even when adequate evidence has been collected. The lack of self-awareness leads to severe \textbf{over-search}, incurring substantial inference latency and prohibitive computational cost. To this end, we propose SAAS, a novel RL framework designed to cultivate dynamic self-awareness that precisely regulates search behavior without compromising accuracy. SAAS introduces three key components: (i) a search boundary modeling mechanism, which identifies the search boundary under the evolving policy by contrasting search-disabled and search-enabled rollouts; (ii) a boundary-aware reward module, which translates this boundary awareness into trajectory-level penalties, suppressing unnecessary and redundant searches; and (iii) a stage-wise optimization strategy, which leverages a sequential curriculum to prioritize reasoning over search regularization, thereby avoiding reward hacking. Extensive experiments demonstrate that SAAS substantially reduces over-search, while maintaining accuracy.
Our code and implementation details are released at \textcolor{blue}{\url{https://github.com/XMUDeepLIT/SAAS}}.

\end{abstract}

\input{section/introduction}
\input{section/preliminary}
\input{section/method}
\input{section/experiment}
\input{section/discussion}

\input{section/limitation}

\bibliography{custom}


\input{section/appendix}

\end{document}

%% file: section/introduction.tex
\section{Introduction}

\begin{figure}[h]
    \centering
    \includegraphics[width=0.50\textwidth]{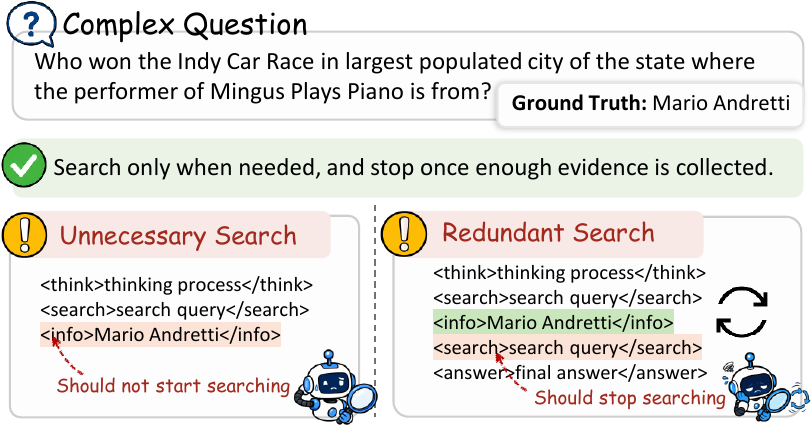}
    \vspace{-6mm}
    \caption{Illustration of two types of over-search in agentic search. \texttt{Unnecessary search} denotes triggering search despite sufficient parametric knowledge, while \texttt{redundant search} denotes continuing search after sufficient external evidence has already been collected.}
\vspace{-3mm}
    \label{fig:intro}

\end{figure}

Large language models (LLMs)~\cite{guo2025deepseek, yang2025qwen3, jaech2024openai} exhibit strong reasoning capabilities across complex tasks, but their reliance on static parametric knowledge can lead to hallucinations in knowledge-intensive settings~\cite{ji2023survey, kandpal2023large}. Agentic search addresses this limitation by coupling reasoning with iterative retrieval, allowing the model to decide when and what to search~\cite{singh2026agenticretrievalaugmentedgenerationsurvey,li2025survey,zhang2026landscapeagenticreinforcementlearning}. Through this reasoning-retrieval cycle, the model decomposes complex questions, retrieves external evidence, and integrates it into subsequent reasoning.

Despite its strengths, agentic search often suffers from critical over-search problem: As shown in Figure~\ref{fig:intro}, the model may trigger unnecessary searches or continue searching even after sufficient evidence is collected~\cite{wu2025searchwiselymitigatingsuboptimal,wu2026hipraghierarchicalprocessrewards}. The former increases unnecessary dependence on external tools, while the latter leads to redundant multi-turn retrieval. Both cases increase computation cost and inference latency, and may introduce noisy evidence that distracts the final answer.

Existing work on over-search can be broadly categorized into two classes: (i) Prompt-based methods that regulate search behavior through prompting strategies or external routing mechanisms without modifying model parameters. Typically, DRAGIN~\cite{su2024dragin} optimizes query formulation to prevent irrelevant exploration, while Adaptive-RAG~\cite{jeong2024adaptive} employs a trained lightweight classifier to assess query complexity and dynamically route requests to the suitable retrieval strategy (ii) RL-based methods which apply reinforcement learning to constrain search depth and restrict excessive tool usage. Typically, StepSearch~\cite{wang2025stepsearchignitingllmssearch} mitigates redundant exploration through planning, decomposing complex question answering into step-wise reasoning and training the model to dynamically schedule search actions at each step, while HiPRAG~\cite{wu2026hipraghierarchicalprocessrewards} assigns fine-grained process rewards based on the informativeness of each retrieved passage, penalizing unnecessary search steps.

\begin{figure}[t]
    \centering
    \includegraphics[width=0.50\textwidth]{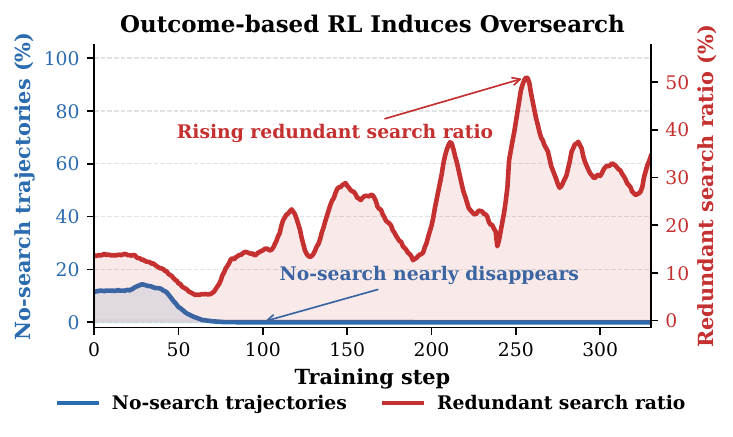}
    \vspace{-8mm}
    \caption{Outcome-based RL induces over-search during training.
The ratio of no-search trajectories (\textcolor{curveblue}{blue}) quickly drops to nearly zero, while the redundant search ratio (\textcolor{curvered}{red}) continues to rise, indicating that RL encourages frequent search and increases redundant search.}
    \label{fig:pre_21}
    \vspace{-5mm}
\end{figure}
However, both paradigms share a common limitation: they rely on static heuristics or fixed thresholds that fail to capture the agent's evolving capability, making them vulnerable to reward hacking, where agents learn to game the penalty rather than develop genuine search boundary awareness. Specifically, existing methods suffer from three critical limitations in practice: (i) outcome-only rewards provide no signal on search awareness, leaving the agent blind to when to initiate and terminate search properly; (ii) naively adding search penalties triggers reward hacking, as static penalties cannot distinguish disciplined restraint from lazy guessing; and (iii) the agent's knowledge distribution changes throughout training, resulting in dynamic shifts in search boundary, yet existing methods provide no mechanism to track these shifts. Without this dynamic self-awareness, the agent is prone to severe over-search, incurring substantial inference latency and prohibitive computational cost.

To this end, we propose SAAS, a \underline{\textbf{S}}elf-\underline{\textbf{A}}ware Reinforcement learning framework for \underline{\textbf{A}}gentic \underline{\textbf{S}}earch, designed to reduce over-search without compromising answer accuracy. It aims to address two fundamental questions: (i) how to dynamically model agent's search boundary as its capability evolves, and (ii) how to translate this awareness into trajectory-level penalties for RL optimization, suppressing unnecessary search without triggering reward hacking. Specifically, SAAS consists of three key components: (i) a search boundary modeling mechanism that contrasts search-disabled and search-enabled rollouts to track the evolving knowledge boundary; (ii) a boundary-aware reward module that translates this boundary awareness into trajectory-level adaptive penalties, suppressing unnecessary search without triggering reward hacking; (iii) a stage-wise optimization strategy that prioritizes deep exploration before search regularization to further stabilize training.

Our major contributions are listed as follows:
\begin{itemize}[leftmargin=*]
    \item We identify the key limitation of existing agentic search models and propose SAAS, a novel reinforcement learning framework that dynamically models the agent's evolving search boundary to suppress unnecessary and redundant searches.

    \item SAAS establishes search boundary awareness over the agent's evolving capability and leverages this awareness to penalizes over-search behavior, combined with a stage-wise policy optimization to prevent reward hacking.

    \item Extensive experiments on seven benchmarks show that SAAS consistently mitigate over-search without compromising model accuracy.

\end{itemize}

%% file: section/preliminary.tex
\section{Preliminary Analysis}
\label{sec:prelim}

In this study, we present a comprehensive analysis of the over-search problem in agentic search from the perspective of optimization dynamics. First, we analyze how standard outcome-based RL optimization drives excessive reliance on external evidence (\S~\ref{sec:prelim1}). We then investigate whether naive search constraints can mitigate this issue and find that fixed penalties fail to reliably define a proper search boundary under an evolving policy (\S~\ref{sec:prelim2}). Finally, we discuss the empirical results in \S~\ref{sec:prelim3} and conclude that \textit{a boundary-aware training framework is needed to overcome this over-search issue}.

\subsection{Over-search Issue in Agentic Search}\label{sec:prelim1}

We first examine how over-search emerges in agentic search. Specifically, we train an agentic search model from scratch with the commonly used outcome-based reward. As shown in Figure~\ref{fig:pre_21}, the model becomes increasingly reliant on search as training proceeds. Since external search often improves final-answer accuracy on complex questions, outcome-based optimization encourages search use but provides no signal about whether a search is necessary or still useful. Specifically, we observe two forms of over-search:

\textbf{Question-level over-search.}
At the question level, the ratio of no-search trajectories steadily decreases during training and becomes nearly zero by step 50. This indicates that outcome-based reward optimization quickly encourages the model to rely on search, making search as its default behavior. Even when internal parametric knowledge is sufficient, the model still performs unnecessary searches. \textit{This suggests that the model fails to learn when search is actually needed.}

\textbf{Step-level redundancy.}
While question-level over-search concerns whether a question requires search, step-level redundancy concerns whether the model can stop searching during reasoning. We measure this using the redundant search ratio, defined as the proportion of searches issued after sufficient evidence has been searched. As shown in Figure~\ref{fig:pre_21}, this ratio rises throughout training and eventually approaches about 50\%, indicating that the model increasingly continues searching even when additional search is no longer needed. These extra searches add little useful information while increasing inference cost and latency. \textit{This suggests that the model fails to judge evidence sufficiency, leading to substantial redundant searches.}

\subsection{Challenges in Learning Search Boundary}\label{sec:prelim2}

The above observations show that outcome-based rewards can induce severe over-search. A straightforward remedy is to impose explicit constraints on search actions, such as directly penalize search calls beyond a predefined threshold. We therefore further conduct a preliminary study on whether such strategy can equip the model with reliable search boundary awareness. Our analysis shows that this remains challenging: not only does the agent's internal knowledge distribution evolve during training, leading to a shifting search boundary, but fixed constraints may also induce reward hacking by discouraging necessary searches. The results are summarized as follows.

\begin{figure}[t]
    \centering
    \includegraphics[width=0.48\textwidth]{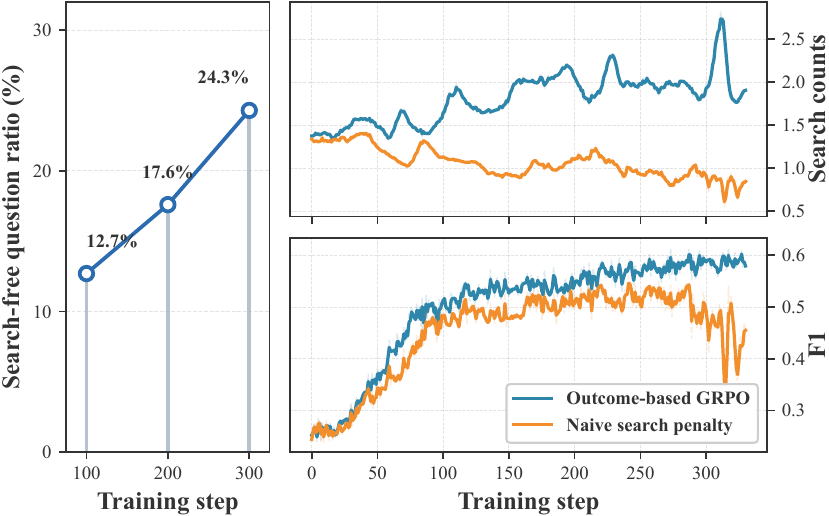}
    \vspace{-6mm}
    \caption{Limitations of naive search penalization. As training progresses, more questions become answerable without search, indicating a shifting search boundary. A naive search penalty reduces search actions but degrades performance, causing late-stage optimization collapse.}
    \vspace{-5mm}
    \label{fig:pre_22}
\end{figure}

\textbf{Search boundary shifts with the model's evolving capability.} Figure~\ref{fig:pre_22} shows that the ratio of questions answerable without search steadily increases from 12.7\% at step 100 to 24.3\% at step 300. This trend indicates that the search boundary is not an intrinsic property of a question, but is conditioned on the model's current capability. As training progresses, questions that initially depend on external search may become solvable with parametric knowledge alone. Therefore, predefined search constraints can become misaligned with the evolving policy, providing unstable or even misleading optimization signals.

\begin{figure*}[t]
    \centering
    \includegraphics[width=\textwidth]{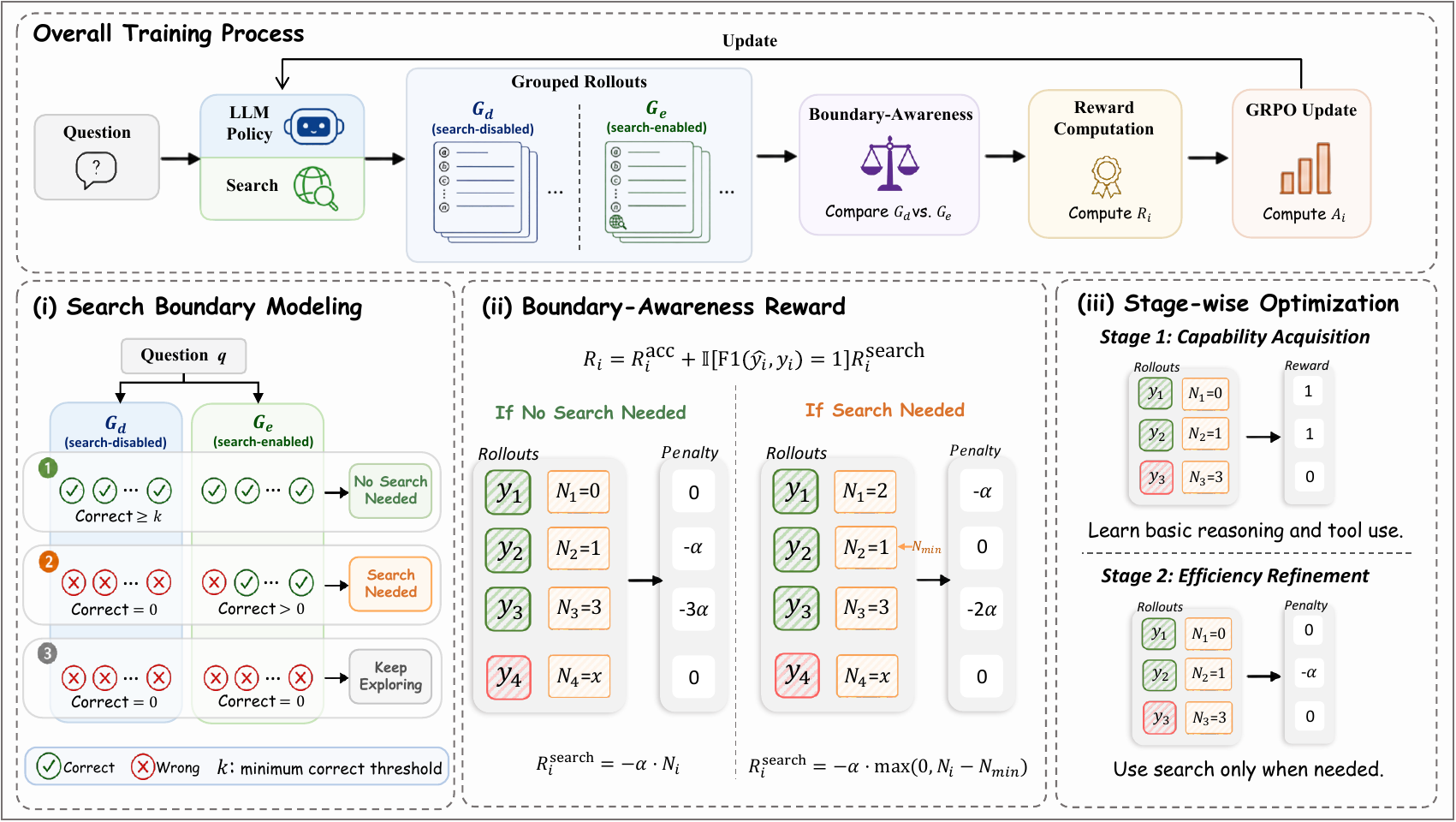}
\caption{
\textbf{The Overall Pipeline of SAAS}. SAAS reduces over-search by training a agent to recognize when search is needed and when further search should stop, organizing optimization around search boundary awareness.
It consists of three components:
(i) \textbf{Search Boundary Modeling} that tracks evolving search boundary by contrasting search-disabled and search-enabled rollouts; (ii) \textbf{Boundary-aware Reward} that translates this boundary awareness into trajectory-level penalties; (iii) \textbf{Stage-wise Optimization} which leverages a sequential curriculum to prioritize deep exploration over search regularization, thereby preventing reward hacking caused by excessive search suppression. 
}
\label{fig:framework}
\vspace{-3mm}
\end{figure*}

\textbf{Fixed penalty can induce reward hacking.}\label{sec:preliminary_fix} As shown in Figure~\ref{fig:pre_22}, applying a fixed penalty to search calls yields lower accuracy than outcome-based training, with training collapse around step 250. The sharp decline in search calls further suggests that the performance drop mainly stems from excessive search suppression. This indicates that naive search penalization does not simply remove redundant searches, but can instead distort the optimization objective. Under a uniform penalty, the model may avoid necessary searches regardless of whether external evidence is still required, resulting in reward hacking and degraded task performance.

\subsection{Discussion}\label{sec:prelim3}

Overall, these preliminary results reveal two limitations of vanilla RL. First, naive search penalties provide no explicit signal for modeling the agent's search boundary, making it hard to decide when search is necessary or existing evidence is sufficient. Second, fixed constraints struggle to balance search efficiency and task performance, as uniform penalties may suppress necessary searches and destabilize optimization. These findings motivate a framework that dynamically models search boundaries for RL optimization to reduce over-search without compromising accuracy.

%% file: section/method.tex
\section{Method}

In this section, we present SAAS, a training framework that reduces over-search by teaching a search-augmented agent both when to search and when to terminate further searches. Specifically, SAAS consists of three key components as shown in Figure~\ref{fig:framework}: (i) \textbf{Search boundary modeling}, which contrasts search-disabled and search-enabled rollouts under the evolving policy to identify each question's search boundary; (ii) \textbf{Boundary-aware reward}, which translates this boundary awareness into trajectory-level penalties, guiding the agent to discriminatively suppress unnecessary and redundant search steps; and (iii) \textbf{Stage-wise optimization}, which leverages a sequential curriculum to prioritize reasoning over search regularization, thereby preventing reward hacking caused by excessive search suppression.

\subsection{Search Boundary Modeling}

For each question, the agent’s search boundary is dynamic: as the policy improves during RL training, questions that initially require external evidence may later become solvable using the model's parametric knowledge alone. Motivated by this observation, SAAS models the search boundary under the evolving policy rather than relying on static annotations or predefined heuristic rules.

Specifically, SAAS identifies the search boundary for each question $q$ by comparing a search-disabled rollout group $G_{\mathrm{d}}(q)$ with a search-enabled rollout group $G_{\mathrm{e}}(q)$ under the evolving policy:
\begin{equation}
\small
\begin{aligned}
G_{\mathrm{d}}(q) &= \{\tau_i \sim \pi_\theta(\cdot \mid q)\}_{i=1}^{N_d}, \\
G_{\mathrm{e}}(q) &= \{\tau_i \sim \pi_\theta(\cdot \mid q, \mathcal{C})\}_{i=1}^{N_e}.
\end{aligned}
\end{equation}
where $\mathcal{C}$ denotes the external evidence searched from the knowledge base during rollout phase. In the search-disabled group $G_{\mathrm{d}}$, the agent is restricted to reasoning solely with its own parametric knowledge, whereas in the search-enabled group $G_{\mathrm{e}}$, it is allowed to interact with the search engine to acquire additional evidence for reasoning.
We then compare the two groups by evaluating the success number of trajectories that produce correct final answers. This comparison reveals whether $q$ lies within the search boundary of the evolving policy. Formally:

\begin{equation}
\small
\begin{aligned}
n_{\mathrm{d}}(q)
&= \sum_{\tau \in G_{\mathrm{d}}(q)}
\mathbb{I}[r_{\mathrm{ans}}(\tau)=1], \\
n_{\mathrm{e}}(q)
&= \sum_{\tau \in G_{\mathrm{e}}(q)}
\mathbb{I}[r_{\mathrm{ans}}(\tau)=1],
\end{aligned}
\end{equation}
where $n_{\mathrm{d}}(q)$ and $n_{\mathrm{e}}(q)$ denote the numbers of correct trajectories in the search-disabled and search-enabled groups. $r_{\mathrm{ans}}(\tau)$ indicates whether trajectory $\tau$ produces the correct final answer. We then define the search boundary of this question as:

\begin{equation}
\small
\mathcal{S}(q)=
\begin{cases}
\textsc{NoSearch}, & n_{\mathrm{d}}(q) \ge \delta, \\
\textsc{NeedSearch}, & n_{\mathrm{d}}(q)=0,\ n_{\mathrm{e}}(q)>0, \\
\textsc{Undetermined}, & \text{otherwise}.
\end{cases}
\end{equation}
where $\delta$ is a threshold that determines whether parametric knowledge is enough to answer the question.

This categorization determines how search behavior should be regulated for $q$. \textsc{NoSearch} indicates that the current policy can reliably solve $q$ without external evidence, while \textsc{NeedSearch} indicates questions that require further searches for successful reasoning. \textsc{Undetermined} covers cases where grouped rollouts provide insufficient evidence to determine the current search boundary.

\subsection{Boundary-Aware Reward}

We next translate the identified search boundary into trajectory-level rewards. Because a uniform penalty over all searches fails to account for the varying search reliance across questions, we design discriminative, boundary-aware rewards that selectively penalize unnecessary and redundant search.

For trajectory $\tau_i$, we define the total reward as:
\begin{equation}
\small
    R_i = R_i^{\mathrm{acc}} + \mathbb{I}[\mathrm{F1}(\hat{y}_i, y_i)=1] R_i^{\mathrm{search}},
\end{equation}
where $R_i^{\mathrm{acc}}$ measures answer quality and $R_i^{\mathrm{search}}$ guides search behavior.
The indicator function $\mathbb{I}$ ensures that the search reward is activated only when the trajectory produces a fully correct answer. We use the F1 score as the accuracy reward:
\begin{equation}
\small
    R_i^{\mathrm{acc}} = \mathrm{F1}(\hat{y}_i, y_i),
\end{equation}
where $\hat{y}_i$ and $y_i$ denote the predicted and golden answers, respectively. Compared with binary correctness metrics, the F1 score provides a smoother accuracy signal. This indicator further prevents excessive suppression of tool use before the model has learned to leverage search effectively.

We instantiate the search reward $R_i^{\mathrm{search}}$ according to the boundary-aware category $\mathcal{S}(q)$. Let $N_i$ denote the number of search actions in trajectory $\tau_i$. For \textsc{NoSearch} questions,  we apply a zero-tolerance penalty:
\begin{equation}
\small
R_i^{\mathrm{search}} = -\alpha N_i.
\end{equation}
For \textsc{NeedSearch} questions, search itself should not be penalized. Instead, to target only redundant searches, we introduce $N_{\min}$ as the minimum sufficient search count, estimated as the fewest search actions among correct trajectories in the search-enabled rollout group:
\begin{equation}
\small
N_{\min} = \min_{\tau_j \in G_{\mathrm{ena}},\, r_a(\tau_j)=1} N_j,
\end{equation}
where $r_a(\tau_j)$ indicates whether $\tau_j$ produces the correct final answer. We then penalize only the search actions that exceed $N_{\min}$:
\begin{equation}
\small
R_i^{\mathrm{search}} = -\alpha \max(0, N_i - N_{\min}).
\end{equation}
For \textsc{Undetermined} questions, the boundary is unclear, so we impose no additional search constraints to avoid restricting reasoning and tool use.

To handle the heterogeneous reward scales between search-disabled and search-enabled rollout groups, we further employ group-wise advantage normalization in GRPO: within each group, advantages are computed and normalized independently. This avoids gradient contamination from distributional mismatch and greatly stabilizes training.

\begin{table*}[htbp]
\centering
\small
\renewcommand{\arraystretch}{1.1}
\setlength{\tabcolsep}{2pt}
\resizebox{\textwidth}{!}{%
\begin{tabular}{l *{16}{c}}
\toprule
\multirow{2}{*}{\textbf{Method}} & \multicolumn{2}{c}{\textbf{TriviaQA}} & \multicolumn{2}{c}{\textbf{PopQA}} & \multicolumn{2}{c}{\textbf{NQ}} & \multicolumn{2}{c}{\textbf{HotpotQA}} & \multicolumn{2}{c}{\textbf{2wiki.}} & \multicolumn{2}{c}{\textbf{Musique}} & \multicolumn{2}{c}{\textbf{Bamboogle}} & \multicolumn{2}{c}{\textbf{AVG}} \\
\cmidrule(lr){2-3} \cmidrule(lr){4-5} \cmidrule(lr){6-7} \cmidrule(lr){8-9} \cmidrule(lr){10-11} \cmidrule(lr){12-13} \cmidrule(lr){14-15} \cmidrule(lr){16-17}
 & ACC & SC & ACC & SC & ACC & SC & ACC & SC & ACC & SC & ACC & SC & ACC & SC & ACC & SC \\
\midrule

\multicolumn{17}{c}{\cellcolor[HTML]{EFEFEF}\textbf{Qwen2.5-3B-Instruct}} \\
\midrule
Direct Inference & 44.9 & - & 14.4 & - & 21.6 & - & 23.6 & - & 22.0 & - & 6.1 & - & 28.8 & - & 23.1 & - \\
RFT~\cite{ahn2024largelanguagemodelsmathematical} & 59.8 & \underline{0.89} & 42.3 & 1.81 & 40.6 & \underline{1.20} & 45.0 & 1.65 & 38.6 & \underline{1.70} & \underline{17.0} & 2.76 & \underline{40.8} & \underline{1.40} & 40.6 & 1.63 \\
Search-R1~\cite{jin2025searchr1trainingllmsreason} & 65.7 & 1.22 & 42.4 & \underline{1.23} & \underline{45.0} & \underline{1.20} & 43.6 & \underline{1.53} & 36.5 & 1.76 & 15.2 & \underline{1.79} & \underline{40.8} & 1.53 & 41.2 & \underline{1.47} \\
StepSearch~\cite{wang2025stepsearchignitingllmssearch} & 60.0 & 1.44 & 39.6 & 1.37 & 41.2 & 1.37 & 45.0 & 1.83 & \underline{41.1} & 2.04 & 16.8 & 2.10 & 27.2 & 1.70 & 38.7 & 1.69 \\
HiPRAG~\cite{wu2026hipraghierarchicalprocessrewards} & \underline{68.8} & 1.59 & \underline{43.8} & 1.63 & \textbf{52.5} & 1.39 & \underline{48.6} & 1.96 & 38.0 & 2.22 & \underline{17.0} & 2.37 & 36.8 & 2.08 & \underline{43.6} & 1.89 \\
\rowcolor{rowblue} \textbf{Ours} & \textbf{69.2} & \textbf{0.66} & \textbf{45.1} & \textbf{1.01} & 43.6 & \textbf{0.72} & \textbf{52.9} & \textbf{1.16} & \textbf{43.9} & \textbf{1.44} & \textbf{20.9} & \textbf{1.62} & \textbf{44.8} & \textbf{1.23} & \textbf{45.8} & \textbf{1.13} \\
\midrule

\multicolumn{17}{c}{\cellcolor[HTML]{EFEFEF}\textbf{Qwen2.5-7B-Instruct}} \\
\midrule
Direct Inference & 55.7 & - & 16.9 & - & 28.2 & - & 28.3 & - & 25.7 & - & 8.8 & - & 40.0 & - & 29.1 & - \\
RFT~\cite{ahn2024largelanguagemodelsmathematical} & 67.5 & \underline{0.92} & \underline{47.3} & 1.46 & 46.2 & 1.19 & 50.1 & 1.64 & 43.2 & 1.75 & 23.1 & 2.50 & 42.4 & 1.36 & 45.7 & 1.56 \\
Search-R1~\cite{jin2025searchr1trainingllmsreason} & \underline{68.3} & 1.11 & 44.9 & 1.26 & 45.6 & 1.24 & 45.8 & \underline{1.19} & 38.7 & \underline{1.43} & 16.8 & \underline{1.38} & 40.0 & \underline{1.15} & 42.9 & \underline{1.25} \\
StepSearch~\cite{wang2025stepsearchignitingllmssearch} & 67.8 & 1.28 & 43.6 & \underline{1.17} & 47.7 & \underline{1.17} & 53.2 & 1.80 & \underline{45.1} & 2.27 & \textbf{26.2} & 2.27 & \textbf{49.6} & 1.84 & 47.6 & 1.69 \\
HiPRAG~\cite{wu2026hipraghierarchicalprocessrewards} & \textbf{74.0} & 2.04 & 46.8 & 2.03 & \textbf{56.0} & 1.96 & \textbf{57.3} & 2.16 & 43.5 & 2.39 & \underline{23.8} & 2.55 & \underline{47.2} & 2.23 & \textbf{49.8} & 2.19 \\
\rowcolor{rowblue} \textbf{SAAS (Ours)} & \textbf{74.0} & \textbf{0.56} & \textbf{47.4} & \textbf{1.01} & \underline{47.8} & \textbf{0.74} & \underline{53.6} & \textbf{0.96} & \textbf{45.9} & \textbf{1.18} & 22.6 & \textbf{1.30} & \textbf{49.6} & \textbf{1.02} & \underline{48.7} & \textbf{0.97} \\

\bottomrule
\end{tabular}%
}

\caption{Performance comparison on seven QA benchmarks. ACC denotes answer accuracy (\%), and SC denotes average search count per question. Best results are in \textbf{bold}, and second-best results are \underline{underlined}.}
\vspace{-5mm}
\label{tab:main_results}
\end{table*}

\subsection{Stage-Wise Optimization}
Although the boundary-aware reward regulates search behavior, activating it too early may induce reward hacking, where the agent avoids necessary searches before acquiring sufficient tool-use competence. We therefore adopt a stage-wise optimization strategy to protect early tool-use learning.

Specifically, the optimization is divided into two stages: capability acquisition and efficiency refinement. In Stage I, the agent is trained only with the outcome-based reward, so it can first learn basic reasoning and tool use without search constraints. When validation performance stops improving, we switch to Stage II and activate the boundary-aware reward to reduce unnecessary and redundant search while preserving answer quality. The trajectory reward is defined as:
\begin{equation}
\small
R_i =
\begin{cases}
R_i^{\mathrm{acc}}, & \text{Stage I}, \\
R_i^{\mathrm{acc}} + \mathbb{I}[\mathrm{F1}(\hat{y}_i, y_i)=1] R_i^{\mathrm{search}}, & \text{Stage II}.
\end{cases}
\end{equation}

In essence, stage-wise optimization leverages a sequential curriculum to prioritize deep exploration over search regularization,
ensuring that the agent first masters when and how to search before it is taught when to refrain. This sequential curriculum mitigates reward hacking and yields a stable training procedure that reduces over-search without compromising agent performance.

%% file: section/experiment.tex
\section{Experiments}

In this section, we aim to answer the following questions. \textbf{Q1} (Main Results): How does SAAS compare with existing baselines in terms of answer accuracy and search efficiency? \textbf{Q2} (Model Analysis):  How does SAAS reduce question-level and step-level over-search, and how do accuracy and search count evolve across training stages? \textbf{Q3} (Ablation Study): How does each component affect accuracy and search efficiency? \textbf{Q4} (Hyperparameter Analysis): How do key hyperparameters affect the performance of SAAS? Please refer to Case Study (\textbf{Q5}) and Efficiency Analysis (\textbf{Q6}) in Appendix~\ref{app:additional_experiments}.

\begin{table*}[htbp]
\centering
\small
\renewcommand{\arraystretch}{1.1}
\setlength{\tabcolsep}{2pt}
\resizebox{\textwidth}{!}{%
\begin{tabular}{l *{16}{c}}
\toprule
\multirow{2}{*}{\textbf{Method}} & \multicolumn{2}{c}{\textbf{TriviaQA}} & \multicolumn{2}{c}{\textbf{PopQA}} & \multicolumn{2}{c}{\textbf{NQ}} & \multicolumn{2}{c}{\textbf{HotpotQA}} & \multicolumn{2}{c}{\textbf{2wiki.}} & \multicolumn{2}{c}{\textbf{Musique}} & \multicolumn{2}{c}{\textbf{Bamboogle}} & \multicolumn{2}{c}{\textbf{AVG}} \\
\cmidrule(lr){2-3} \cmidrule(lr){4-5} \cmidrule(lr){6-7} \cmidrule(lr){8-9} \cmidrule(lr){10-11} \cmidrule(lr){12-13} \cmidrule(lr){14-15} \cmidrule(lr){16-17}
 & SOR & QOR & SOR & QOR & SOR & QOR & SOR & QOR & SOR & QOR & SOR & QOR & SOR & QOR & SOR & QOR \\
\midrule

RFT~\cite{ahn2024largelanguagemodelsmathematical} 
& \underline{22.8} & \underline{40.9}
& \underline{7.6} & \underline{59.3}
& \underline{15.6} & \underline{61.2}
& \underline{12.6} & \textbf{40.0}
& \underline{5.8} & \underline{42.0}
& \underline{12.7} & 79.7
& 26.5 & \underline{44.4}
& \underline{14.8} & \underline{52.5} \\

Search-R1~\cite{jin2025searchr1trainingllmsreason}           
& 37.8 & 91.2
& 12.6 & 94.2
& 24.1 & 97.9
& 14.8 & 81.2
& 4.8 & 75.5
& \underline{12.7} & \textbf{75.0}
& \underline{25.2} & 64.3
& 18.9 & 82.8 \\

StepSearch~\cite{wang2025stepsearchignitingllmssearch}          
& 37.2 & 99.8
& 15.5 & 99.9
& 26.8 & 99.8
& 20.0 & 99.9
& 6.8 & 100.0
& 19.9 & 100.0
& 43.7 & 100.0
& 24.3 & 99.9 \\

HiPRAG~\cite{wu2026hipraghierarchicalprocessrewards}             
& 30.8 & 100.0
& 12.4 & 100.0
& 21.2 & 100.0
& 15.6 & 100.0
& \underline{4.2} & 100.0
& 15.1 & 100.0
& 36.9 & 100.0
& 19.5 & 100.0 \\

\rowcolor{rowblue} \textbf{Ours} 
& \textbf{11.5} & \textbf{29.4} 
& \textbf{2.1} & \textbf{45.8}
& \textbf{8.7} & \textbf{52.6}
& \textbf{4.1} & \underline{40.7}
& \textbf{1.7} & \textbf{36.9}
& \textbf{2.7} & \underline{78.0}
& \textbf{13.3} & \textbf{38.1} 
& \textbf{6.3} & \textbf{45.9} \\

\bottomrule
\end{tabular}%
}
\caption{Over-search analysis on Qwen2.5-7B-Instruct. SOR and QOR denote step-level and question-level over-search ratios (\%), respectively. A lower SOR indicates that the model performs fewer redundant searches, while a lower QOR indicates that the model better leverages its parametric knowledge to avoid unnecessary search.}
\label{tab:oversearch-analysis}
\vspace{-5mm}
\end{table*}

\subsection{Experimental Setup}

\paragraph{Datasets.}
We evaluate SAAS on seven open-domain QA benchmarks, covering both single-hop and multi-hop scenarios. The single-hop setting includes  TriviaQA~\cite{joshi2017triviaqalargescaledistantly}, PopQA~\cite{mallen2023trustlanguagemodelsinvestigating} and NQ~\cite{kwiatkowski2019natural}. The multi-hop setting includes HotpotQA~\cite{yang2018hotpotqadatasetdiverseexplainable}, 2WikiMultiHopQA~\cite{ho2020constructingmultihopqadataset}, MuSiQue~\cite{trivedi2022musiquemultihopquestionssinglehop}, and Bamboogle~\cite{press2023measuringnarrowingcompositionalitygap}. 
More details on datasets are provided in Appendix~\ref{app:benchmark_dataset}.

\paragraph{Baselines.}
We compare SAAS against a comprehensive set of baselines
that represent different paradigms in agentic search: (1) \textbf{Direct Inference}: direct generation without any search mechanism. (2) \textbf{Rejection Sampling Fine Tuning (RFT)}~\cite{ahn2024largelanguagemodelsmathematical}: fine-tuning the model on trajectories generated through rejection sampling. (3) \textbf{RL-Based Agentic Search}: use reinforcement learning to train agentic search capacity, including Search-R1~\cite{jin2025searchr1trainingllmsreason}, StepSearch~\cite{wang2025stepsearchignitingllmssearch}, and HiPRAG~\cite{wu2026hipraghierarchicalprocessrewards}. Please refer to more details in Appendix~\ref{app:baseline_details}. 

\paragraph{Evaluation Metrics.}
We evaluate each method from two aspects: answer quality and search behavior. For answer quality, we report \textbf{Accuracy (Acc)}, measured by an LLM judge following prior work~\cite{chen2025researchlearningreasonsearch}. For search behavior, we report \textbf{Search Count (SC)}, \textbf{Question-level Over-search Ratio (QOR)}, and \textbf{Step-level Over-search Ratio (SOR)}. SC measures the average number of search calls, QOR measures the ratio of unnecessary searches on questions answerable from parametric knowledge, while SOR measures the ratio of redundant search actions. Detailed metric definitions are provided in Appendix~\ref{app:evaluation_details}. 

\paragraph{Implementation Details.}
We construct the training corpus by combining the training splits of NQ and HotpotQA, and employ E5~\cite{wang2024textembeddingsweaklysupervisedcontrastive} as the dense encoder to embed queries for evidence search over the 2018 Wikipedia dump~\cite{karpukhin2020densepassageretrievalopendomain}. We use GPT-4 as the LLM judge for answer evaluation. Various LLMs serve as backbone models for the experiment. More implementation details are included in Appendix~\ref{app:sysname_setup}.

\subsection{Main Results (Q1)}

To address Q1, we evaluate SAAS with direct inference, RFT, and other RL-based agentic search baselines 
across seven open-domain benchmarks. According to Table~\ref{tab:main_results} which reports accuracy and search count on Qwen2.5 series model, we summarize the key observations below.

\textbf{SAAS consistently outperforms baselines.}
SAAS achieves the best average accuracy on Qwen2.5-3B-Instruct ($45.8\%$), surpassing the strongest baseline HiPRAG by $2.2\%$, and remains competitive on Qwen2.5-7B-Instruct ($48.7\%$). The gain is especially clear on multi-hop tasks. Specifically, SAAS improves over HiPRAG by $8.0\%$ on Bamboogle. This suggests that modeling search boundaries does not compromise reasoning quality, but helps focus search on questions where external evidence is truly needed.

\textbf{SAAS substantially reduces redundant search.}
SAAS consistently uses fewer searches than baselines, reducing the average search count to $1.13$ on Qwen2.5-3B-Instruct and $0.97$ on Qwen2.5-7B-Instruct. In contrast, strong baselines such as StepSearch and HiPRAG require $1.69$ and $2.19$ searches on average, respectively. This shows that SAAS avoids default search behavior while preserving necessary retrieval for reasoning.

\textbf{SAAS achieves a better accuracy-efficiency trade-off.} SAAS achieves a favorable balance between answer accuracy and search cost. Specifically, on Qwen2.5-3B-Instruct, it improves accuracy over the strongest baseline from $43.6\%$ to $45.8\%$ while using $40.2\%$ fewer search calls. This indicates that SAAS does not simply reduce search frequency, but learns a clearer search boundary that better aligns retrieval with actual evidence needs.

\subsection{Model Analysis (Q2)}
\subsubsection{Over-search Behavior Analysis}

To verify whether SAAS directly mitigates over-search, we further measure two over-search-related metrics: QOR and SOR. QOR captures whether the model triggers search on questions that can be answered without search, while SOR captures whether it issues redundant searches during the reasoning process. The results on Qwen2.5-7B-Instruct are reported in Table~\ref{tab:oversearch-analysis}.

\textbf{SAAS effectively reduces question-level over-search.}
As shown in Table~\ref{tab:oversearch-analysis}, SAAS achieves the lowest average QOR ($45.9\%$), clearly lower than RFT ($52.5\%$) and much lower than search-heavy baselines such as StepSearch ($99.9\%$) and HiPRAG ($100.0\%$). This shows that SAAS better identifies when the model's parametric knowledge is sufficient and avoids unnecessary searches.

\begin{figure}[t]
    \centering
    \includegraphics[width=\linewidth]{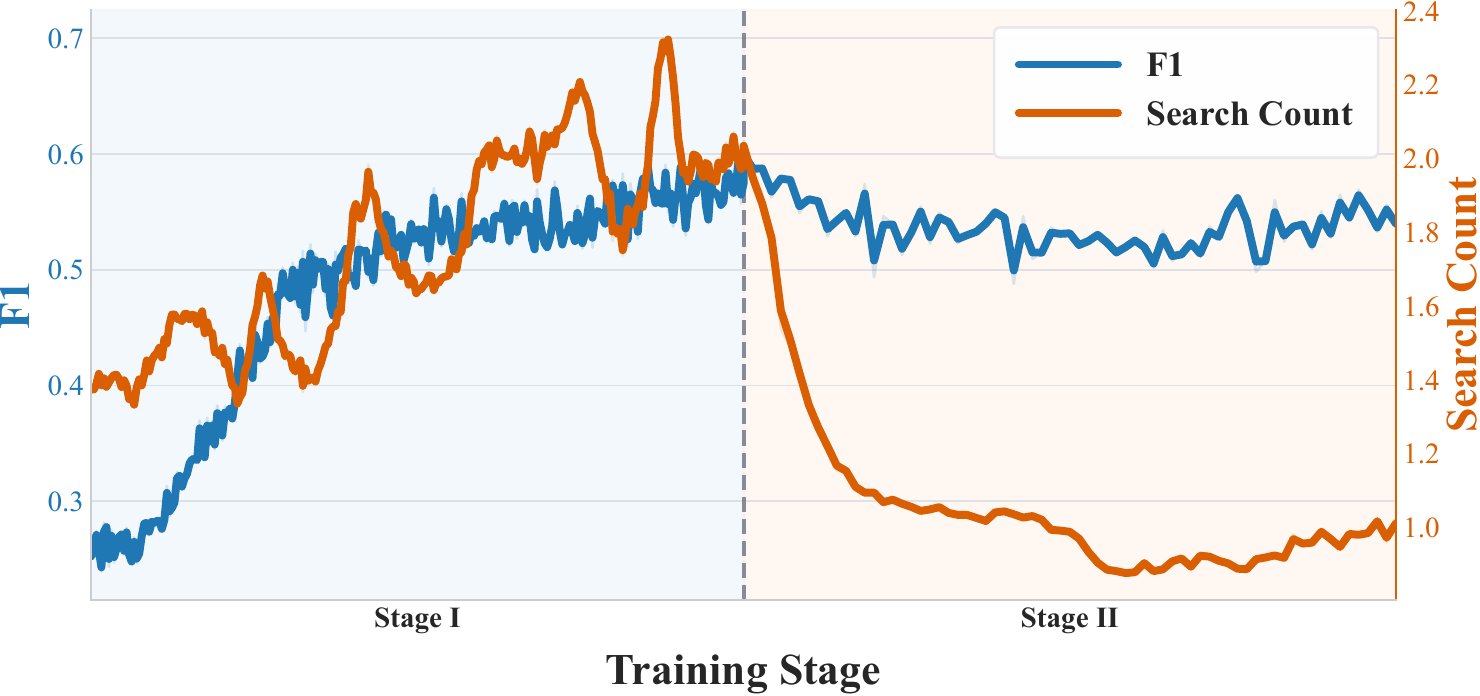}
    \caption{Training dynamics of F1 and search count across the two-stage RL training process.}
    \label{fig:training_dynamics}
    \vspace{-5mm}
\end{figure}

\textbf{SAAS also suppresses step-level redundant search.}
SAAS achieves the lowest SOR on every benchmark, reducing the average SOR to $6.3\%$, far below RFT ($14.8\%$) and StepSearch ($24.3\%$). The reduction is also clear on multi-hop datasets, such as 2WikiMultiHopQA, where SOR drops to $1.7\%$. These results show that SAAS learns to stop once sufficient external evidence has been collected, encouraged by the $N_{\min}$-based reward that penalizes searches beyond the sufficient-evidence boundary.

\subsubsection{Training Dynamics}

To better understand how our Boundary-Awareness Guided Reward affects search-augmented agent training, we track the training dynamics of average F1 and search count across two stages on Qwen2.5-3B-Instruct. The results are shown in Figure~\ref{fig:training_dynamics}.

Figure~\ref{fig:training_dynamics} shows two distinct training stages. In Stage I, before the boundary-aware reward is introduced, both average F1 and search count increase, indicating that the model first learns to leverage the search tool to improve answer quality. In Stage II, after activating the boundary-aware reward, the average search count drops sharply from about $2.0$ to below $1.0$, while F1 shows only a mild temporary decrease and then remains stable. Unlike the naive penalty setting in Section~\ref{sec:prelim2}, this transition does not cause training collapse. Instead, SAAS recalibrates the search boundary, reducing over-search behavior while preserving answer quality.

\begin{figure}[t]
    \centering
    \includegraphics[width=\linewidth]{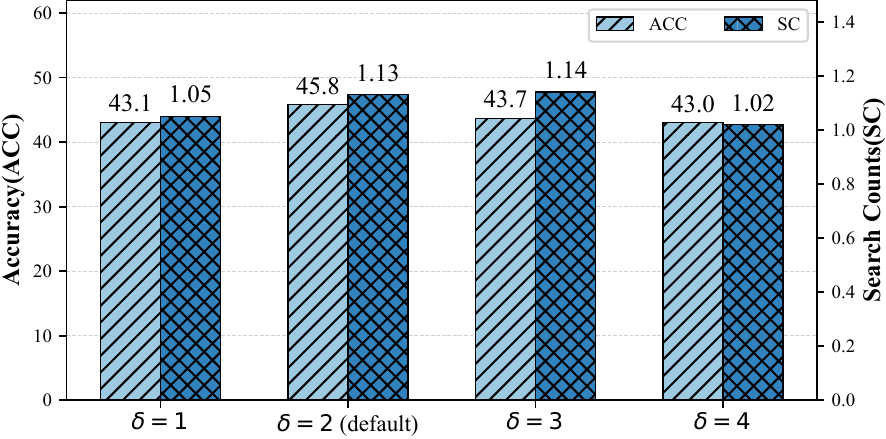}
    \vspace{-5mm}
    \caption{Sensitivity analysis of the key hyperparameter $\delta$, the number of GRPO groups used for evidence-demand estimation. ACC and RC are averaged across seven QA benchmarks on Qwen2.5-3B-Instruct.}
    \label{fig:app_delta_sensitivity}
    \vspace{-5mm}
\end{figure}

\subsection{Ablation Study (Q3)}

To analyze the contribution of each component in SAAS, we conduct ablation studies on Qwen2.5-3B-Instruct. As shown in Table~\ref{tab:ablation_results}, SAAS achieves the best balance between  accuracy and search count, indicating that stage-wise optimization and on-policy boundary modeling jointly contribute to stable and boundary-aware search behavior.

\textbf{Stage-wise optimization promotes stable search learning.} As shown in Table~\ref{tab:ablation_results}, removing stage-wise optimization further reduces the average search count, but substantially lowers accuracy from $45.8\%$ to $40.9\%$. This suggests that optimizing for search efficiency from the beginning can over-constrain the policy before it acquires reliable tool-use behavior. In contrast, stage-wise optimization first develops the agent's tool-use capability and then refines search efficiency, yielding a better accuracy-efficiency trade-off.

\textbf{On-policy boundary modeling better matches the evolving policy.}
Without on-policy boundary modeling, the search boundary is derived from the base model before training and then kept fixed during optimization. This variant achieves only $42.8\%$ accuracy with $1.07$ search count, below the full method. This shows that the search boundary shifts as the model improves. A boundary fixed before training can become misaligned with the current policy, while on-policy boundary modeling keeps the search-regulation signal actively updated.

\subsection{Parameter Sensitivity Analysis (Q4)}
We further conduct a sensitivity analysis on the thershold hyperparameter $\delta$ used in search boundary modeling. Specifically, we vary $\delta$ on Qwen2.5-3B-Instruct and evaluate the resulting average answer accuracy and search count across the seven QA benchmarks. This experiment examines whether the default setting provides a stable search-boundary modeling for reward assignment. More detailed analysis are provided in Appendix~\ref{app:parameter_sensitivity}.

\textbf{The sensitivity analysis confirms that $\delta=2$ provides the most reliable search-boundary signal for downstream policy optimization.} Under this setting, SAAS achieves the highest average ACC of 45.8\% with a low average SC of 1.13, indicating a favorable accuracy-efficiency balance. Reducing the threshold to $\delta=1$ slightly lowers SC to 1.05, but ACC drops to 43.1\%, showing that the lower search count comes from over-suppressing search behavior rather than yielding more reliable search-boundary awareness. Increasing the threshold to $\delta=3$ or $\delta=4$ also degrades ACC to 43.7\% and 43.0\%, respectively, suggesting that stricter grouping introduces less stable boundary modeling and weakens the reward signal for necessary evidence acquisition. To summarize, overly aggressive compression of search actions leads to accuracy degradation, whereas $\delta=2$ preserves sufficient evidence acquisition while discouraging unnecessary and redundant search steps.

\begin{table}[t]
\centering
\small
\renewcommand{\arraystretch}{1.1}
\setlength{\tabcolsep}{12.2pt}
\begin{tabular}{lcc}
\toprule
\textbf{Method Variant} & \textbf{ACC} & \textbf{SC} \\
\midrule
\rowcolor{rowblue} \textbf{SAAS} & 45.8 & 1.13 \\
w/o Stage-wise optimization & 40.9 & 0.95  \\
w/o On-policy estimation & 42.8 & 1.07 \\
\bottomrule
\end{tabular}
\caption{Ablation study on SAAS components using Qwen2.5-3B-Instruct as the backbone language model. 
}
\label{tab:ablation_results}
\vspace{-5mm}
\end{table}

%% file: section/discussion.tex
\section{Conclusion}

In this paper, we introduce SAAS, a self-aware reinforcement learning framework for mitigating over-search in agentic search. SAAS models the evolving search boundary of the RL policy and integrates boundary-aware rewards with stage-wise optimization to regulate search behavior. Experiments across 7 open-domain benchmarks show that SAAS effectively reduces both question-level and step-level over-search while maintaining strong performance. 
These results underscore the importance of dynamic search boundary awareness for more efficient and effective search-augmented reasoning.

%% file: section/limitation.tex
\clearpage
\section*{Limitation}

While SAAS effectively regulates search behavior within text-based agentic systems, our current evaluation focuses on unimodal textual retrieval. In practice, knowledge-intensive tasks may also involve multimodal evidence, such as images, tables, and structured databases. Although textual serialization can capture the essential semantics of these sources, directly incorporating raw multimodal signals may provide richer contextual information. Importantly, the core formulation of SAAS is inherently agnostic to input modalities, as its fundamental mechanism of modeling the boundary of required information to prevent over-search does not rely on text-specific features. Consequently, generalizing this framework to multimodal reasoning constitutes a natural progression for future research. Such an extension would enable the model to perform cross-modal reasoning, grounding textual information in visual, structural, or other multimodal evidence, thereby moving toward more comprehensive search-augmented reasoning systems.

\section*{Ethics Statement}

We confirm that this study follows the Ethics Policy. 
All data and models used are publicly available and contain no private user information.

\paragraph{Data and Model Bias}
Our experiments use publicly available open-domain QA benchmarks and a Wikipedia-based corpus. We do not collect private user data. These datasets are well-established and widely adopted in the research community, with no known systematic biases that would affect the validity of our conclusions. The pre-trained language models used in our experiments are standard off-the-shelf models, while our method, SARS, focuses solely on regulating search behavior based on the model’s evolving knowledge boundary, and does not rely on or amplify any sensitive attributes.

\paragraph{Intended Use}
SAAS is designed as a research framework to support researchers and developers in studying efficient search-augmented reasoning. By reducing redundant search, SAAS aims to improve the deployability of search-augmented agents in practical scenarios where inference latency and retrieval cost are important considerations.

%% file: section/appendix.tex
\appendix

\section{Frequently Asked Questions (FAQs)}

\subsection{Code and Data Availability}

To facilitate future research and allow independent verification of our results, all code and data associated with SAAS are released via an anonymous repository: \textcolor{blue}{\url{https://anonymous.4open.science/r/SAAS-50B0}}. This repository provides a comprehensive experimental framework. Specifically, it details our exact methodology, including the rollout generation process, reward calculations, and the precise stage-wise reinforcement learning setups. Coupled with the release of all processed datasets, standardized prompts, search interface templates, and full evaluation scripts, this repository grants researchers unrestricted access to every component required to reproduce our work.

\subsection{What are the advantages of SAAS?}
SAAS presents several advantages over existing agentic search methods by explicitly modeling the agent's evolving search boundary. This search boundary awareness improves search efficiency while preserving the ability to acquire external evidence when search is genuinely needed:

\noindent\textbf{Reduction of unnecessary search.} Many questions can be answered using the model's parametric knowledge alone, yet RL-based search agents trained with outcome-only rewards may still trigger search because unnecessary tool use is not explicitly penalized. SAAS addresses this issue by contrasting search-disabled and search-enabled rollouts to identify low-reliance questions where the model's parametric knowledge is sufficient. For these questions, search actions are penalized, encouraging the model to avoid unnecessary search at the question level.

\noindent\textbf{Suppression of redundant search.} For questions that require external evidence, the key challenge is not only whether to search, but also when to stop searching. Outcome-only rewards can encourage the model to continue issuing search calls even after sufficient external evidence has been collected. SAAS uses the minimum successful search trajectory as a reference and penalizes searches beyond this sufficient-evidence boundary. As a result, it suppresses redundant search without discouraging evidence acquisition when search is necessary.

\noindent\textbf{Stable accuracy-efficiency trade-off.} SAAS does not simply minimize search count. Instead, it adopts stage-wise optimization to first develop basic reasoning and tool-use capability, and then applies boundary-aware reward assignment to refine search behavior. This design reduces search count while maintaining competitive answer accuracy across single-hop and multi-hop benchmarks as well as different model backbones.

\subsection{Why not simply penalize every search action?}
A uniform penalty on all search actions is too coarse because the usefulness of search actions depends on the question and the current policy. For questions that can be answered using the model's parametric knowledge alone, any search action is unnecessary and should be discouraged. In contrast, for questions that require external evidence under the current policy, search is useful until sufficient evidence has been collected, after which additional search calls become redundant. Treating these cases identically can therefore suppress necessary evidence acquisition and harm answer accuracy, while still providing no guidance on when the model should initiate or terminate search. SAAS addresses this limitation through boundary-aware reward assignment based on the estimated search necessity of the current policy. When a question does not require search, SAAS penalizes search actions to reduce question-level over-search. When search is required, SAAS allows evidence acquisition but penalizes searches beyond the minimum successful search trajectory, thereby suppressing step-level redundant search. This design reduces both unnecessary and redundant searches while preserving tool-use ability when external evidence is genuinely needed.

\begin{table*}[t]
\centering
\small
\renewcommand{\arraystretch}{1.1}
\resizebox{\textwidth}{!}{%
\begin{tabular}{l *{16}{c}}
\toprule
\multirow{2}{*}{\textbf{Method}} 
& \multicolumn{2}{c}{\textbf{TriviaQA}} 
& \multicolumn{2}{c}{\textbf{PopQA}} 
& \multicolumn{2}{c}{\textbf{NQ}} 
& \multicolumn{2}{c}{\textbf{HotpotQA}} 
& \multicolumn{2}{c}{\textbf{2wiki.}} 
& \multicolumn{2}{c}{\textbf{Musique}} 
& \multicolumn{2}{c}{\textbf{Bamboogle}} 
& \multicolumn{2}{c}{\textbf{AVG}} \\
\cmidrule(lr){2-3} \cmidrule(lr){4-5}
\cmidrule(lr){6-7} \cmidrule(lr){8-9}
\cmidrule(lr){10-11} \cmidrule(lr){12-13}
\cmidrule(lr){14-15} \cmidrule(lr){16-17}
& ACC & SC & ACC & SC & ACC & SC & ACC & SC & ACC & SC & ACC & SC & ACC & SC & ACC & SC \\
\midrule

\multicolumn{17}{c}{\cellcolor[HTML]{EFEFEF}\textbf{Qwen2.5-3B-Instruct}} \\
\midrule
GRPO
& 69.3 & 1.87 
& 46.3 & 1.91 
& 46.4 & 1.86 
& 53.7 & 2.30 
& 45.4 & 2.63 
& 21.8 & 2.88 
& 42.4 & 2.34 
& 46.5 & 2.26 \\
\rowcolor{rowblue} \textbf{Ours} 
& 69.2 & 0.66 
& 45.1 & 1.01 
& 43.6 & 0.72 
& 52.9 & 1.16 
& 43.9 & 1.44 
& 20.9 & 1.62 
& 44.8 & 1.23 
& 45.8 & 1.13 \\

\midrule
\multicolumn{17}{c}{\cellcolor[HTML]{EFEFEF}\textbf{Qwen2.5-7B-Instruct}} \\
\midrule
GRPO
& 73.9 & 2.26 
& 48.0 & 2.40 
& 48.3 & 2.32 
& 57.6 & 3.02 
& 49.2 & 3.86 
& 25.3 & 3.63 
& 48.8 & 3.10 
& 50.2 & 2.94 \\
\rowcolor{rowblue} \textbf{Ours} 
& 74.0 & 0.56 
& 47.4 & 1.01 
& 47.8 & 0.74 
& 53.6 & 0.96 
& 45.9 & 1.18 
& 22.6 & 1.30 
& 49.6 & 1.02 
& 48.7 & 0.97 \\

\midrule
\multicolumn{17}{c}{\cellcolor[HTML]{EFEFEF}\textbf{Qwen3-4B-Instruct}} \\
\midrule
GRPO
& 74.6 & 1.70 
& 49.4 & 1.65 
& 49.5 & 1.70 
& 61.7 & 2.37 
& 50.2 & 2.89 
& 29.0 & 2.83 
& 60.8 & 2.35 
& 53.6 & 2.21 \\
\rowcolor{rowblue} \textbf{Ours} 
& 73.6 & 0.69 
& 49.0 & 1.19 
& 49.6 & 0.91 
& 58.2 & 1.25 
& 53.3 & 1.79 
& 27.3 & 1.72 
& 58.4 & 1.20 
& 52.8 & 1.25 \\

\bottomrule
\end{tabular}%
}
\caption{Efficiency Analysis. Accuracy and search count comparison between outcome-based GRPO and SAAS across three backbones. ACC denotes answer accuracy (\%), and SC denotes average search count per question.}
\label{tab:app_grpo_ours_acc_rc}
\end{table*}

\subsection{Why is the search boundary modeled on-policy?}
The search boundary is policy-dependent and evolves throughout RL training. As the model improves, questions that initially require external evidence may become solvable using parametric knowledge alone. For questions that still require search, the boundary for terminating search can also shift, since the policy may learn to acquire sufficient evidence with fewer search actions. A boundary derived only from the base model or from fixed heuristic rules can therefore become misaligned with the current policy. SAAS avoids this mismatch by modeling the search boundary on-policy: it contrasts search-disabled and search-enabled rollout groups from the current model to determine whether search should be initiated and how much search is sufficient. This keeps the search-regulation signal updated during training and enables boundary-aware reward assignment to suppress both question-level unnecessary search and step-level redundant search. The ablation without on-policy boundary modeling performs worse than the full method, further confirming the importance of tracking the evolving search boundary.

\subsection{How does SAAS differ from prior efficient agentic search methods?}
Prior efficient agentic search methods typically regulate search behavior through heuristic triggers, confidence signals, search-depth constraints, or process rewards. Although these mechanisms can reduce search frequency, they do not explicitly align search decisions with the model's evolving search boundary: whether the current policy can answer from parametric knowledge alone, and whether the acquired external evidence is already sufficient. Unlike these approaches, SAAS dynamically tracks the agent's actual competence during training. By anchoring the reward mechanism directly to the policy's evolving state, it discriminatively evaluates the necessity of search for each specific question. This ensures that the agent is penalized for over-search only when search initiation is strictly unnecessary or when the gathered evidence is already sufficient, fundamentally differing from the static constraints used in previous methods.

\section{Additional Experiments}\label{app:additional_experiments}

\subsection{SAAS Setup}\label{app:sysname_setup}

During training, we use grouped rollouts to estimate the search label $\mathcal{S}(q)$ and compute the search reward. For each question, the policy samples $N=8$ trajectories, which are evenly split into two groups: four search-disabled rollouts and four search-enabled rollouts. The search-disabled group estimates whether the question can be solved from the model's parametric knowledge, while the search-enabled group estimates whether search can help solve difficult questions. Following the search-label definition in the method section, we set the grouping threshold to $\delta=2$: a question is labeled as \textsc{NoSearch} if at least $\delta$ search-disabled rollouts answer it correctly, and as \textsc{NeedSearch} if no search-disabled rollout succeeds but at least one search-enabled rollout is correct.

For SAAS-specific reward settings, we set the search penalty coefficient in $R_i^{\mathrm{search}}$ to $\alpha=0.05$. For \textsc{NoSearch} questions, the reward penalizes every search call according to $N_i$; for \textsc{NeedSearch} questions, it penalizes only extra calls beyond $N_{\min}$, the minimum number of searches among correct search-enabled trajectories. During both training and evaluation, each query retrieves the top $k=3$ documents. The maximum number of search calls in a trajectory is set to $5$.

Meanwhile, we implement the GRPO algorithm based on the slime\footnote{https://github.com/THUDM/slime} framework. The detailed hyperparameter settings are listed in Table~\ref{tab:sysname_hypers}.

\subsection{Case Study (Q5)}\label{app:case_study_oversearch}
To qualitatively illustrate the over-search patterns discussed above, we provide two representative case studies in Figure~\ref{fig:case_unnecessary_search} and Figure~\ref{fig:case_redundant_search}. These examples compare SAAS with the GRPO baseline trained only with correctness rewards. The first case focuses on unnecessary search, where the question can be directly answered from parametric knowledge but outcome-based optimization still encourages searches. The second case focuses on redundant search after search has already been triggered: both methods obtain the correct answer, but outcome-based GRPO continues issuing extra queries, whereas SAAS stops earlier once sufficient evidence has been collected.

\begin{table}[h]
\centering
\small
\begin{tcolorbox}[
  width=\columnwidth,
  colframe=black!70,
  colback=gray!8,
  boxrule=0.4pt,
  arc=1pt,
  left=4pt,
  right=4pt,
  top=4pt,
  bottom=4pt
]

\setlength{\tabcolsep}{5pt}
\renewcommand{\arraystretch}{0.92}

\begin{tabular}{@{}ll@{}}
\toprule
\textbf{Hyperparameter} & \textbf{Value} \\
\midrule
Learning Rate / Scheduler 
& $1\times10^{-6}$ / Constant \\

Warmup Ratio / Epochs 
& $0.285$ / $1$ \\

Batch Size / Global Batch Size 
& $512$ / $4096$ \\

Rollouts / Rollout Temp. 
& $8$ / $1.0$ \\

KL Coeff. / Clip Ratio ($\epsilon$) 
& $0.001$ / $0.2$ \\

Max Prompt / Response Len. 
& $512$ / $512$ \\
\bottomrule
\end{tabular}

\end{tcolorbox}
\caption{Training hyperparameter settings of SAAS.}
\label{tab:sysname_hypers}
\end{table}

\subsection{Efficiency Analysis (Q6)}\label{app:perf_eff_tradeoff}
Table~\ref{tab:app_grpo_ours_acc_rc} illustrates the accuracy-efficiency trade-off of SAAS compared to a standard GRPO baseline across seven QA benchmarks and three model backbones. Effectiveness is quantified by answer accuracy, while efficiency is tracked via the average search count. To establish a rigorous comparison, the GRPO baseline is trained solely with an outcome-based F1 reward.

\begin{table*}[htbp]
\centering
\small
\renewcommand{\arraystretch}{1.1}
\resizebox{\textwidth}{!}{%
\begin{tabular}{l *{16}{c}}
\toprule
\multirow{2}{*}{\textbf{Method}} 
& \multicolumn{2}{c}{\textbf{TriviaQA}} 
& \multicolumn{2}{c}{\textbf{PopQA}} 
& \multicolumn{2}{c}{\textbf{NQ}} 
& \multicolumn{2}{c}{\textbf{HotpotQA}} 
& \multicolumn{2}{c}{\textbf{2wiki.}} 
& \multicolumn{2}{c}{\textbf{Musique}} 
& \multicolumn{2}{c}{\textbf{Bamboogle}} 
& \multicolumn{2}{c}{\textbf{AVG}} \\
\cmidrule(lr){2-3} \cmidrule(lr){4-5}
\cmidrule(lr){6-7} \cmidrule(lr){8-9}
\cmidrule(lr){10-11} \cmidrule(lr){12-13}
\cmidrule(lr){14-15} \cmidrule(lr){16-17}
& SOR & QOR & SOR & QOR & SOR & QOR & SOR & QOR & SOR & QOR & SOR & QOR & SOR & QOR & SOR & QOR \\
\midrule

\multicolumn{17}{c}{\cellcolor[HTML]{EFEFEF}\textbf{Qwen2.5-3B-Instruct}} \\
\midrule
GRPO 
& 26.4 & 100.0
& 10.4 & 100.0
& 16.5 & 100.0
& 11.6 & 100.0
& 5.0 & 100.0
& 11.1 & 100.0 
& 28.5 & 100.0
& 15.6 & 100.0 \\
\rowcolor{rowblue} \textbf{Ours} 
& 21.2 & 41.6 
& 6.5 & 61.8 
& 10.5 & 46.2 
& 8.6 & 52.2 
& 4.4 & 73.7 
& 6.6 & 82.9 
& 15.7 & 65.6 
& 10.5 & 60.6 \\

\midrule
\multicolumn{17}{c}{\cellcolor[HTML]{EFEFEF}\textbf{Qwen2.5-7B-Instruct}} \\
\midrule
GRPO
& 30.1 & 100.0
& 12.5 & 100.0
& 18.4 & 100.0
& 12.1 & 100.0
& 3.7 & 100.0
& 8.1 & 100.0
& 22.6 & 100.0
& 15.4 & 100.0 \\
\rowcolor{rowblue} \textbf{Ours} 
& 11.5 & 29.4 
& 2.1 & 45.8
& 8.7 & 52.6
& 4.1 & 40.7
& 1.7 & 36.9
& 2.7 & 78.0
& 13.3 & 38.1 
& 6.3 & 45.9 \\

\midrule
\multicolumn{17}{c}{\cellcolor[HTML]{EFEFEF}\textbf{Qwen3-4B-Instruct}} \\
\midrule
GRPO
& 25.4 & 100.0 
& 11.4 & 100.0 
& 16.5 & 100.0 
& 10.2 & 100.0
& 2.7 & 100.0
& 6.7 & 100.0
& 20.4 & 100.0
& 13.3 & 100.0 \\
\rowcolor{rowblue} \textbf{Ours} 
& 10.5 & 27.4
& 3.6 & 47.1
& 8.3 & 44.2
& 4.8 & 39.5
& 2.2 & 65.4
& 4.5 & 56.9
& 8.8 & 52.1
& 6.1 & 47.5 \\

\bottomrule
\end{tabular}%
}

\caption{Over-search comparison between outcome-based GRPO and SAAS. SOR and QOR denote step-level and question-level over-search ratios, respectively. Lower values indicate fewer redundant and unnecessary searches.}
\label{tab:app_grpo_ours_sor_qor}
\end{table*}

\textbf{SAAS significantly reducing search overhead without compromising answer quality.} On Qwen2.5-3B-Instruct, SAAS reduces the average RC from 2.26 to 1.13, corresponding to a 50.0\% reduction in search count, while the average ACC decreases only slightly from 46.5\% to 45.8\%. The same trend becomes even more pronounced on Qwen2.5-7B-Instruct: SAAS lowers the average RC from 2.94 to 0.97, reducing by about 67.0\%, while retaining a competitive average ACC of 48.7\% compared with 50.2\% for GRPO. On Qwen3-4B-Instruct, SAAS also reduces the average RC from 2.21 to 1.25, with only a 0.8\% drop in average accuracy. These results indicate that the trade-off between model's performance and efficiency is not tied to a specific backbone, but generalizes across models.

\textbf{Outcome-based GRPO introduces a severe search burden.} As shown in Table~\ref{tab:app_grpo_ours_acc_rc}, GRPO consistently incurs high average search count across all three backbones, reaching 2.26, 2.94, and 2.21 on Qwen2.5-3B-Instruct, Qwen2.5-7B-Instruct, and Qwen3-4B-Instruct, respectively. This observation is consistent with the motivation: when the reward only reflects final answer's correctness, search is treated as a uniformly beneficial action because additional information may increase the chance of matching the gold answer. As a result, GRPO lacks an explicit signal for distinguishing necessary information acquisition from redundant tool use, and thus tends to learn a conservative policy that repeatedly searches even after sufficient information has been obtained. In contrast, SAAS introduces search-necessity aware supervision to couple search decisions with the model's actual knowledge boundary. Search is encouraged when external evidence is needed, but unnecessary or redundant search actions are penalized. This mechanism explains why SAAS can substantially reduce search count while retaining comparable accuracy: it does not simply suppress search, but reallocates search actions to samples and reasoning stages where external evidence is most valuable.

\begin{table*}[t]
    \centering
    \includegraphics[width=0.95\linewidth]{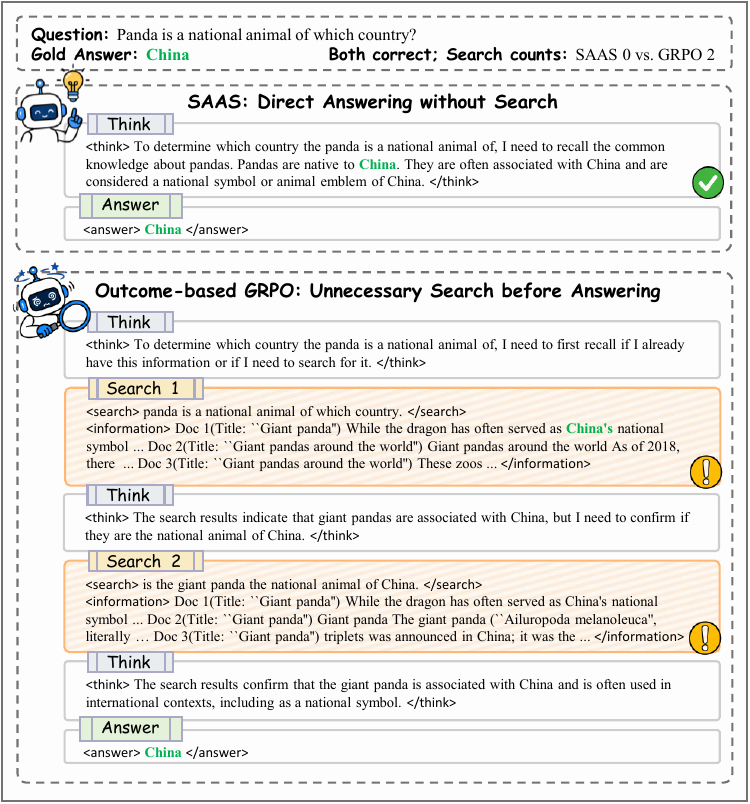}
    \caption{Case study on unnecessary search.The question can be answered using parametric knowledge alone, yet outcome-based GRPO still triggers searches, whereas SAAS answers directly without search.}
  \label{fig:case_unnecessary_search}
\end{table*}

\begin{table*}[htbp]
    \centering
    \includegraphics[width=0.95\linewidth]{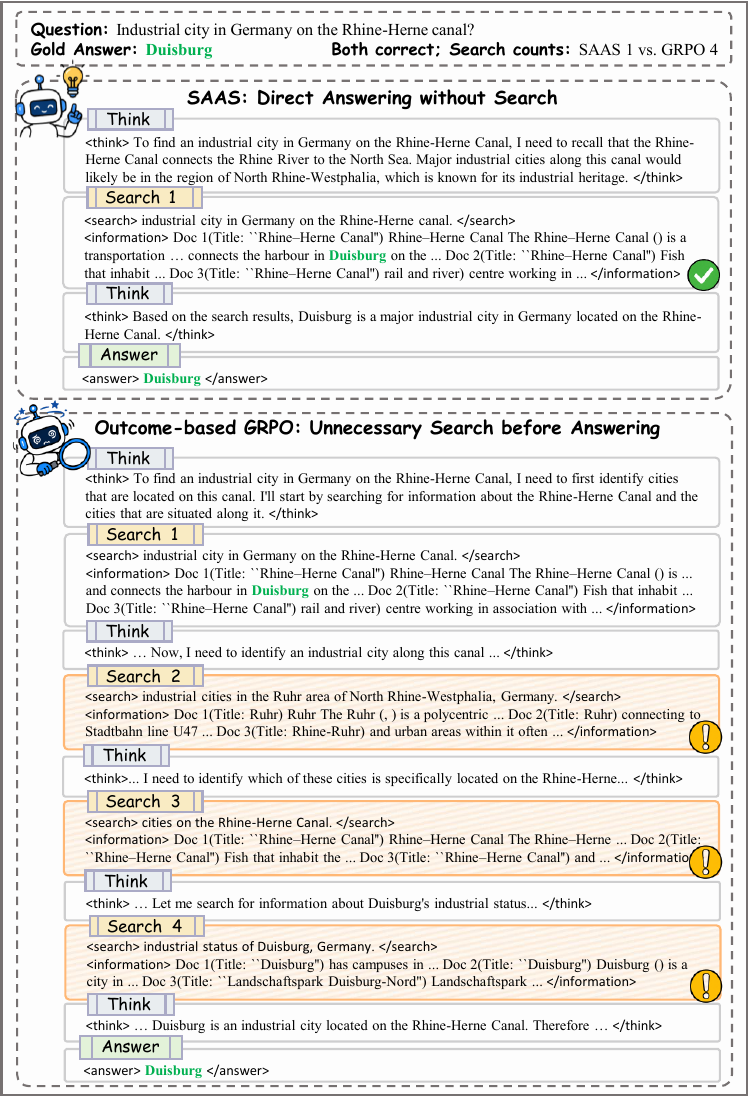}
    \caption{Case study on redundant search behavior. Both methods answer correctly, but outcome-based GRPO repeatedly issues additional queries before producing the final answer, whereas SAAS reaches the answer with substantially fewer search calls.}
  \label{fig:case_redundant_search}

\end{table*}

\subsection{Parameter Sensitivity Analysis}\label{app:parameter_sensitivity}
Finally, we analyze the sensitivity of the grouping hyperparameter $\delta$ used for GRPO-grouped search-necessity estimation. Figure~\ref{fig:app_delta_sensitivity} reports the average ACC and RC under different values of $\delta$ on Qwen2.5-3B-Instruct, aiming to verify whether the default setting provides reliable necessity estimates for training.

\textbf{The default setting $\delta=2$ yields the most effective necessity estimation for downstream policy learning.} With $\delta=2$, SAAS achieves the highest average ACC of 45.8\% while maintaining a low average RC of 1.13. When $\delta=1$, the average RC remains comparable at 1.05, but the average ACC drops to 43.1\%. This indicates that overly coarse grouping can still suppress searches, but it provides less accurate search-necessity estimates and therefore weakens the reward signal used to distinguish samples that require external information from those that do not.

\textbf{Using more groups does not improve training effectiveness because finer grouping may introduce noisier necessity estimates.} When $\delta=3$ and $\delta=4$, the average ACC decreases to 43.7\% and 43.0\%, respectively, while the average RC remains close to the default setting. These results suggest that increasing $\delta$ does not produce a more useful supervision signal; instead, excessively fine grouping may make the estimated evidence demand less stable, leading to less consistent search decisions during training. Overall, the sensitivity analysis supports the choice of $\delta=2$, which provides the best balance between accurate necessity estimation and stable policy optimization.

\textbf{These efficiency gains are particularly pronounced on tasks prone to excessive searches.} Multi-hop benchmarks such as 2WikiMultihopQA and MuSiQue require the model to gather information across multiple reasoning steps, which creates more opportunities for unnecessary tool calls once the relevant information has already been obtained. Under the GRPO baseline, the outcome-based F1 reward does not distinguish necessary information acquisition from redundant searches, so the model tends to keep querying in these complex settings, reaching 3.86 and 3.63 search count on 2WikiMultihopQA and MuSiQue on Qwen2.5-7B-Instruct. By explicitly regularizing search necessity, SAAS learns to stop searching when additional information provides limited marginal value, reducing the corresponding search count to 1.18 and 1.30.

\subsection{Over-search Behavior Compared with Outcome-Based GRPO}\label{app:oversearch_behavior}
Table~\ref{tab:app_grpo_ours_sor_qor} further analyzes the over-search behavior of SAAS and the outcome-based GRPO baseline. We report SOR and QOR to evaluate two complementary aspects of search overhead: SOR measures redundant search actions after retrieval has been initiated, while QOR measures unnecessary search on questions that can be answered from the model's parametric knowledge.

\textbf{SAAS substantially reduces unnecessary searches across model backbones.} On Qwen2.5-3B-Instruct, GRPO yields an average QOR of 100.0\%, indicating that it triggers search for all questions answerable from the model's parametric knowledge. In contrast, SAAS reduces the average QOR to 60.6\%, corresponding to a 39.4\% reduction in question-level over-search. The improvement becomes more pronounced on Qwen2.5-7B-Instruct, where the average QOR decreases from 100.0\% to 45.9\%. On Qwen3-4B-Instruct, GRPO still exhibits a QOR of 100.0 across all datasets, whereas SAAS reduces the average QOR to 47.5\%. These results show that SAAS consistently suppresses unnecessary search actions, rather than simply reducing search on a specific backbone.

\textbf{SAAS also mitigates redundant search after search has been triggered.} Across all three backbones, SAAS consistently lowers the average SOR compared with GRPO. On Qwen2.5-3B-Instruct, the average SOR decreases from 15.6\% to 10.5\%. On Qwen2.5-7B-Instruct, the reduction is more substantial, from 15.4\% to 6.3\%, with particularly large improvements on TriviaQA, PopQA, HotpotQA, and MuSiQue. The same trend also holds for Qwen3-4B-Instruct, where the average SOR drops from 13.3\% to 6.1\%. These results indicate that the proposed method not only learns whether searches should be invoked, but also improves the model's ability to terminate search once sufficient information has been collected.

\textbf{The reduction in both QOR and SOR demonstrates that SAAS addresses over-search at different decision levels.} Question-level improvements show that the model avoids initiating search when external information is unnecessary, while step-level improvements indicate fewer redundant searches after search begins. This behavior is consistent with the design of SAAS, which encourages search only when it is expected to provide meaningful additional information. Consequently, SAAS achieves more targeted tool use and alleviates the over-search tendency induced by outcome-only optimization.

\begin{table*}[htbp]

\begin{tcolorbox}[colback=gray!10, colframe=black!100, title=LLM-as-a-Judge Evaluation, width=\textwidth]
You are an evaluation assistant. Please determine if the model output is equivalent to the labeled answer.
\\
Question: \{question\}
\\
Labeled Answer: \{labeled answer\}
\\
Model Output: \{pred answer\}
\\
Did the model give an answer equivalent to the labeled answer? Please respond with "Correct" if they are equivalent, or "Incorrect" if they are not equivalent. Do not include any other text.
\end{tcolorbox}
\label{tab:llm-prompt}

\begin{tcolorbox}[colback=gray!10, colframe=black!100, title=SOR Semantic Consistency Evaluation, width=\textwidth]
You are an expert in Semantic Analysis. Evaluate whether Response 1 and Response 2 provide the same core answer to the given Search Query.
\\
Output \texttt{True} if their answers are semantically consistent, otherwise output \texttt{False}.
\\
\textless Inputs\textgreater
\\
Search Query: \{query\}
\\
Response 1: \{response1\}
\\
Response 2: \{response2\}
\\
\textless /Inputs\textgreater
\\
Provide your output as a single boolean value strictly enclosed in tags.
\\
Example: \texttt{\textless answer\textgreater True\textless /answer\textgreater}
\end{tcolorbox}

\begin{tcolorbox}[colback=gray!10, colframe=black!100, title=Search-Enabled QA Prompt, width=\textwidth]
Answer the given question. You must conduct reasoning inside \texttt{\textless think\textgreater} and \texttt{\textless /think\textgreater} first every time you get new information. After reasoning, if you find you lack some knowledge, you can call a search engine by \texttt{\textless search\textgreater query \textless /search\textgreater} and it will return the top searched results between \texttt{\textless information\textgreater} and \texttt{\textless /information\textgreater}. You can search as many times as your want. If you find no further external knowledge needed, you can directly provide the answer inside \texttt{\textless answer\textgreater} and \texttt{\textless /answer\textgreater}, without detailed illustrations. For example, \texttt{\textless answer\textgreater Beijing \textless /answer\textgreater}. Question: \{question\}
\end{tcolorbox}

\begin{tcolorbox}[colback=gray!10, colframe=black!100, title=Search-Disabled QA Prompt, width=\textwidth]
Answer the given question. You must conduct reasoning inside \texttt{\textless think\textgreater} and \texttt{\textless /think\textgreater} first. After reasoning, you can directly provide the answer inside \texttt{\textless answer\textgreater} and \texttt{\textless /answer\textgreater}, without detailed illustrations. For example, \texttt{\textless answer\textgreater Beijing \textless /answer\textgreater}. Question: \{question\}
\end{tcolorbox}
\label{app:search_disabled_prompt}

\caption{Prompt templates used for answer evaluation, SOR semantic-consistency evaluation, search-enabled and search-disabled question answering throughout our experiments.}
\label{tab:prompt_set}

\end{table*}

\section{Implementation Details}
\subsection{Benchmark Dataset}\label{app:benchmark_dataset}

To comprehensively evaluate SAAS across questions with different degrees of reliance on search, we conduct experiments on seven open-domain QA benchmarks. The single-hop setting includes TriviaQA, PopQA and Natural Questions (NQ), which generally have lower search boundary because many questions focus on one factual relation and can often be answered using the model's parametric knowledge. In contrast, the multi-hop setting includes HotpotQA, 2WikiMultiHopQA, MuSiQue, and Bamboogle, where questions require the model to gather evidence across reasoning steps, making them more search-dependent and often requiring multiple search actions to answer correctly.

\noindent\textbf{TriviaQA}~\cite{joshi2017triviaqalargescaledistantly}: TriviaQA is built from trivia questions collected from online sources. For each question, it normalizes the reference answers and gathers related evidence documents from Wikipedia and the web, so that the answer can be checked against external textual evidence.

\noindent\textbf{PopQA}~\cite{mallen2023trustlanguagemodelsinvestigating}: PopQA is built from Wikidata triples by converting relations such as occupations, birthplaces, or authorship into natural-language questions. The object in each Wikidata triple is used as the short answer, producing factual questions that test whether a model knows specific entity-relation facts.

\noindent\textbf{Natural Questions (NQ)}~\cite{kwiatkowski2019natural}: NQ comprises real-world anonymized queries issued to the Google search engine. The dataset pairs each query with a corresponding Wikipedia page, and provides human annotations for both long-answer passages and short answer spans.

\noindent\textbf{HotpotQA}~\cite{yang2018hotpotqadatasetdiverseexplainable}: HotpotQA is built from Wikipedia and asks questions that require evidence from multiple pages. Examples are annotated with supporting sentences, making it suitable for evaluating if a model can retrieve and combine evidence for bridge or comparison questions.

\noindent\textbf{2WikiMultiHopQA}~\cite{ho2020constructingmultihopqadataset}: 2WikiMultiHopQA is constructed from Wikidata relations and linked Wikipedia passages. It uses predefined reasoning patterns to generate multi-hop questions and provides annotations for intermediate entities, evidence documents, and the reasoning path needed to reach the answer.

\noindent\textbf{MuSiQue}~\cite{trivedi2022musiquemultihopquestionssinglehop}: MuSiQue is built by composing connected single-hop questions into harder multi-hop questions. This construction reduces shortcut solutions, since the final answer depends on following a sequence of reasoning steps supported by paragraph-level evidence.

\noindent\textbf{Bamboogle}~\cite{press2023measuringnarrowingcompositionalitygap}: Bamboogle is a manually written set of compositional questions designed to require multiple independent facts. Its questions are created to be difficult to answer from a single clue, so models usually need to search for and combine evidence across steps.

\subsection{Baseline Details}\label{app:baseline_details}

In this section, we provide detailed descriptions of each baseline used in our comparison.

\noindent\textbf{Direct Inference}\quad directly prompts the base model to answer each question without the search engine. This baseline evaluates the model's ability to answer from parametric knowledge alone and serves as the reference for measuring the benefit and cost of external search.

\noindent\textbf{Rejection Sampling Fine Tuning}~\cite{ahn2024largelanguagemodelsmathematical}\quad fine-tunes the model on trajectories selected through rejection sampling. For each training question, the model generates one search-enabled trajectory and one search-disabled trajectory; only trajectories that lead to correct answers are retained for fine-tuning. This baseline tests whether learning from successful sampled trajectories is sufficient to acquire useful search behavior without reinforcement learning.

\noindent\textbf{Search-R1}~\cite{jin2025searchr1trainingllmsreason}\quad trains LLMs to reason and interact with a search engine through reinforcement learning. During generation, the model can issue search queries, read retrieved evidence, and continue reasoning before producing the final answer. Its training objective mainly rewards final answer correctness, making it a representative outcome-based RL baseline for agentic search.

\noindent\textbf{StepSearch}~\cite{wang2025stepsearchignitingllmssearch}\quad improves search efficiency from a planning perspective. It decomposes complex question answering into step-wise reasoning and search decisions, and uses proximal policy optimization to train the model to plan when and what to retrieve at each step. By coordinating searches with the evolving reasoning process, StepSearch serves as a planning-based baseline for evaluating whether better search scheduling can reduce inefficient searches.

\noindent\textbf{HiPRAG}~\cite{wu2026hipraghierarchicalprocessrewards}\quad introduces hierarchical process rewards for efficient agentic search-augmented generation. Instead of relying only on final answer correctness, it provides reward signals over search actions and reasoning steps, helping the model improve answer quality and search efficiency. We include HiPRAG as a strong process-reward baseline for comparison with SAAS.

\subsection{Evaluation Metrics}\label{app:evaluation_details}

To comprehensively evaluate each method, we employ four metrics from two aspects: answer quality and search behavior. Answer quality is evaluated using Accuracy (ACC), while search behavior is evaluated using Search Count (SC), Question-level Over-search Ratio (QOR), and Step-level Over-search Ratio (SOR). All prompts used for evaluation are provided in Appendix~\ref{app:prompt}.

\noindent\textbf{Accuracy (ACC)} measures whether the final answer is correct. Following prior work~\cite{chen2025researchlearningreasonsearch, liu2026bapoboundaryawarepolicyoptimization}, we use an LLM judge to determine whether the model output is equivalent to the ground truth. Given a test set $D=\{q_i\}_{i=1}^{|D|}$, ACC is calculated as:
\begin{equation}
\mathrm{ACC} = \frac{1}{|D|}\sum_{i=1}^{|D|}\mathbb{I}\big(\mathrm{Judge}(\hat{y}_i, y_i)=1\big),
\end{equation}
where $\hat{y}_i$ denotes the model prediction for question $q_i$, $y_i$ denotes the ground truth, and $\mathbb{I}(\cdot)$ is the indicator function.

\noindent\textbf{Search Count (SC)} measures the average number of search calls among all trajectories. Let $\tau_i$ denote the generated trajectory for question $q_i$, and let $N_i$ denote the number of search calls in $\tau_i$. SC is defined as:
\begin{equation}
\mathrm{SC} = \frac{1}{|D|}\sum_{i=1}^{|D|} N_i.
\end{equation}
A lower SC indicates that the model uses fewer search calls on average.

\noindent\textbf{Question-level Over-search Ratio (QOR)} measures unnecessary search at the question level. We define $D_{\mathrm{para}}=\{q_i\}_{i=1}^{|D_{\mathrm{para}}|}$ as the subset of questions that can be correctly answered using the model's parametric knowledge without search. For each question $q_i\in D_{\mathrm{para}}$, $o_i=1$ if the generated trajectory $\tau_i$ still contains search actions. QOR is calculated as:
\begin{equation}
\mathrm{QOR} = \frac{\sum_{i=1}^{|D_{\mathrm{para}}|} o_i}{|D_{\mathrm{para}}|}.
\end{equation}
A lower QOR indicates that the model better leverages its parametric knowledge to avoid unnecessary search.

\noindent\textbf{Step-level Over-search Ratio (SOR)} measures redundant search at the step level. Following ~\cite{wu2026hipraghierarchicalprocessrewards}, for each question $q_i\in D$, let $\mathcal{S}_{\tau_i}$ denote the set of search queries in the generated trajectory $\tau_i$. For each search query $s\in\mathcal{S}_{\tau_i}$, $o_s=1$ if the model's answer to $s$ is semantically consistent with the retrieved documents, indicating that this search step is redundant. SOR is defined as:
\begin{equation}
\mathrm{SOR} = \frac{\sum_{i=1}^{|D|}\sum_{s\in\mathcal{S}_{\tau_i}} o_s}{\sum_{i=1}^{|D|}|\mathcal{S}_{\tau_i}|}.
\end{equation}
A lower SOR indicates that the model performs fewer redundant search steps.

\section{Related Work}

\subsection{Agentic Search}

Moving beyond static retrieve-then-generate RAG~\cite{zhang2025survey,gao2025probing,zhuang2025linearrag,xiang2025use,xiao2026reliable,chen2026you,zhang2025faithfulrag}, agentic search studies how models can dynamically interleave reasoning with external search~\cite{singh2026agenticretrievalaugmentedgenerationsurvey,yang2026graph,xiang2026systematic,li2025survey, zhang2026landscapeagenticreinforcementlearning}. ReAct~\cite{yao2023reactsynergizingreasoningacting} introduced the reasoning and acting paradigm, enabling models to decide when to invoke tools. IRCoT~\cite{trivedi2023interleavingretrievalchainofthoughtreasoning} adapted this paradigm to more complex question answering by interleaving retrieval with chain-of-thought reasoning, so that intermediate reasoning steps can be grounded in retrieved evidence. Along this line, Self-RAG~\cite{asai2023selfraglearningretrievegenerate} and FLARE~\cite{jiang2023activeretrievalaugmentedgeneration} introduced reflection tokens and confidence-driven retrieval triggers to make search decisions more adaptive during decoding. Chain-of-Retrieval~\cite{wang2025chainofretrievalaugmentedgeneration} and Search-o1~\cite{li2025searcho1agenticsearchenhancedlarge} further refine this process by supporting dynamic query formulation, while DeepResearcher~\cite{zheng2025deepresearcherscalingdeepresearch} and WebThinker~\cite{li2025webthinkerempoweringlargereasoning} extend agentic search to  web environments that require browsing and evidence synthesis. More recently, reinforcement learning has been adopted to train tool-use policies~\cite{qian2025toolrlrewardtoollearning, feng2025retoolreinforcementlearningstrategic, li2025torlscalingtoolintegratedrl}. Representative systems optimize search-augmented reasoning through task-success rewards, encouraging models to invoke search and exploit retrieved evidence~\cite{jin2025searchr1trainingllmsreason, chen2025researchlearningreasonsearch, song2025r1searcherincentivizingsearchcapability, dao2025rezeroenhancingllmsearch, jiang2025deepretrievalhackingrealsearch}. To reduce the cost and instability of relying on live search engines during training, ZeroSearch~\cite{sun2026zerosearchincentivizesearchcapability} and SSRL~\cite{fan2025ssrlselfsearchreinforcementlearning} simulate search environments, while MaskSearch~\cite{wu2025masksearchuniversalpretrainingframework} uses search-augmented pre-training to elicit generalizable search abilities. However, these outcome-driven approaches often underemphasize search efficiency, leaving models prone to over-search.

\subsection{Efficient Agentic Search}

While agentic search improves complex reasoning, it often introduces unnecessary or redundant tool calls. Existing efforts can be broadly grouped into three directions. One direction improves planning ability to orchestrate when and what to query. DRAGIN~\cite{su2024dragin} dynamically schedules searches based on real-time information needs, while PAR-RAG~\cite{zhang2026credibleplandrivenragmethod} employs complexity-aware top-down planning to prevent irrelevant exploration and redundant retrieval. The second direction investigates retrieval necessity. Early studies~\cite{dhole2025retrieveretrieveuncertaintydetection, jeong2024adaptive} primarily depended on auxiliary classifiers or routers to evaluate query difficulty and guide search. In contrast, more recent methods~\cite{zubkova2025sugarleveragingcontextualconfidence, yao2025seakr, huanshuo2025ctrla, zhang2025kbmdelineatingknowledgeboundary} decode self-awareness signals from the LLM's internal hidden states or explicitly delineate the parametric knowledge boundary to decide when search is truly required. Beyond these directions, reinforcement learning directly optimizes tool-use efficiency. Several approaches incorporate efficiency metrics into rewards to penalize suboptimal tool usage. For example, SEM~\cite{sha2025semreinforcementlearningsearchefficient} trains intrinsically search-efficient models, Search Wisely~\cite{wu2025searchwiselymitigatingsuboptimal} mitigates unnecessary searches by reducing model uncertainty, and IKEA~\cite{huang2025reinforcedinternalexternalknowledgesynergistic} introduces a knowledge-boundary aware reward to prioritize parametric knowledge and resort to search only when internal knowledge is insufficient. Other works focus on adaptive exploration. R1-Searcher++~\cite{song2025r1searcherincentivizingdynamicknowledge} and AutoSearch~\cite{sun2026autosearchadaptivesearchdepth} use RL to incentivize dynamic knowledge acquisition and flexible search depths, while S3~\cite{jiang2025s3dontneeddata} decouples the searcher from a frozen generator and trains it with a Gain-Beyond-RAG reward to improve downstream utility with limited data. Recent frameworks integrate process rewards for finer-grained supervision. ReARTeR~\cite{sun2025rearter} evaluates and refines individual steps using process rewards. HiPRAG~\cite{wu2026hipraghierarchicalprocessrewards} and parallel works on process rewards~\cite{ye2025correctnessharmonizingprocessoutcome, yue2025promotingefficientreasoningverifiable} introduce hierarchical supervision to constrain reasoning paths and curb over-searching. Despite these advances, these methods often lack a unified mechanism to trigger and halt search based on knowledge boundaries.

\section{Prompt Set}\label{app:prompt}

To facilitate reproducibility and provide transparency into the evaluation and agentic retrieval behaviors, we present the specific prompt sets used throughout our experiments in Table~\ref{tab:prompt_set}. These prompts instantiate the answer evaluator, the SOR semantic-consistency evaluator, the search-enabled and search-disabled question-answering settings. Together, they standardize the evaluation protocol and orchestrate the reasoning--search process described in the main text.

\section{The Use of Large Language Models}

In preparing this paper, we made limited use of Large Language Models (LLMs). Specifically, LLMs were used to polish the writing, correct grammatical errors, and improve sentence clarity and readability without altering the scientific content. LLMs were also used to assist with minor formatting tasks, such as improving the presentation of tables and organizing related-work descriptions. All model-assisted suggestions were carefully reviewed and revised by the authors, who retain full responsibility for the scientific accuracy, technical claims, and integrity of the final manuscript.

%% file: custom.bib
@misc{singh2026agenticretrievalaugmentedgenerationsurvey,
      title={Agentic Retrieval-Augmented Generation: A Survey on Agentic RAG}, 
      author={Aditi Singh and Abul Ehtesham and Saket Kumar and Tala Talaei Khoei and Athanasios V. Vasilakos},
      year={2026},
      eprint={2501.09136},
      archivePrefix={arXiv},
      primaryClass={cs.AI},
      url={https://arxiv.org/abs/2501.09136}, 
}

@article{li2025survey,
  title={A Survey on AI Search with Large Language Models},
  author={Li, Jian and Li, Xiaoxi and Zheng, Yan and Jin, Yizhang and Wang, Shuo and Wu, Jiafu and Wang, Yabiao and Wang, Chengjie and Yuan, Xiaotong},
  year={2025}
}

@misc{zhang2026landscapeagenticreinforcementlearning,
      title={The Landscape of Agentic Reinforcement Learning for LLMs: A Survey}, 
      author={Guibin Zhang and Hejia Geng and Xiaohang Yu and Zhenfei Yin and Zaibin Zhang and Zelin Tan and Heng Zhou and Zhongzhi Li and Xiangyuan Xue and Yijiang Li and Yifan Zhou and Yang Chen and Chen Zhang and Yutao Fan and Zihu Wang and Songtao Huang and Francisco Piedrahita-Velez and Yue Liao and Hongru Wang and Mengyue Yang and Heng Ji and Jun Wang and Shuicheng Yan and Philip Torr and Lei Bai},
      year={2026},
      eprint={2509.02547},
      archivePrefix={arXiv},
      primaryClass={cs.AI},
      url={https://arxiv.org/abs/2509.02547}, 
}

@misc{wang2025stepsearchignitingllmssearch,
      title={StepSearch: Igniting LLMs Search Ability via Step-Wise Proximal Policy Optimization}, 
      author={Ziliang Wang and Xuhui Zheng and Kang An and Cijun Ouyang and Jialu Cai and Yuhang Wang and Yichao Wu},
      year={2025},
      eprint={2505.15107},
      archivePrefix={arXiv},
      primaryClass={cs.CL},
      url={https://arxiv.org/abs/2505.15107}, 
}

@misc{yao2023reactsynergizingreasoningacting,
      title={ReAct: Synergizing Reasoning and Acting in Language Models}, 
      author={Shunyu Yao and Jeffrey Zhao and Dian Yu and Nan Du and Izhak Shafran and Karthik Narasimhan and Yuan Cao},
      year={2023},
      eprint={2210.03629},
      archivePrefix={arXiv},
      primaryClass={cs.CL},
      url={https://arxiv.org/abs/2210.03629}, 
}

@misc{trivedi2023interleavingretrievalchainofthoughtreasoning,
      title={Interleaving Retrieval with Chain-of-Thought Reasoning for Knowledge-Intensive Multi-Step Questions}, 
      author={Harsh Trivedi and Niranjan Balasubramanian and Tushar Khot and Ashish Sabharwal},
      year={2023},
      eprint={2212.10509},
      archivePrefix={arXiv},
      primaryClass={cs.CL},
      url={https://arxiv.org/abs/2212.10509}, 
}

@misc{wang2025chainofretrievalaugmentedgeneration,
      title={Chain-of-Retrieval Augmented Generation}, 
      author={Liang Wang and Haonan Chen and Nan Yang and Xiaolong Huang and Zhicheng Dou and Furu Wei},
      year={2025},
      eprint={2501.14342},
      archivePrefix={arXiv},
      primaryClass={cs.IR},
      url={https://arxiv.org/abs/2501.14342}, 
}

@misc{li2025searcho1agenticsearchenhancedlarge,
      title={Search-o1: Agentic Search-Enhanced Large Reasoning Models}, 
      author={Xiaoxi Li and Guanting Dong and Jiajie Jin and Yuyao Zhang and Yujia Zhou and Yutao Zhu and Peitian Zhang and Zhicheng Dou},
      year={2025},
      eprint={2501.05366},
      archivePrefix={arXiv},
      primaryClass={cs.AI},
      url={https://arxiv.org/abs/2501.05366}, 
}

@misc{jin2025searchr1trainingllmsreason,
      title={Search-R1: Training LLMs to Reason and Leverage Search Engines with Reinforcement Learning}, 
      author={Bowen Jin and Hansi Zeng and Zhenrui Yue and Jinsung Yoon and Sercan Arik and Dong Wang and Hamed Zamani and Jiawei Han},
      year={2025},
      eprint={2503.09516},
      archivePrefix={arXiv},
      primaryClass={cs.CL},
      url={https://arxiv.org/abs/2503.09516}, 
}

@misc{chen2025researchlearningreasonsearch,
      title={ReSearch: Learning to Reason with Search for LLMs via Reinforcement Learning}, 
      author={Mingyang Chen and Linzhuang Sun and Tianpeng Li and Haoze Sun and Yijie Zhou and Chenzheng Zhu and Haofen Wang and Jeff Z. Pan and Wen Zhang and Huajun Chen and Fan Yang and Zenan Zhou and Weipeng Chen},
      year={2025},
      eprint={2503.19470},
      archivePrefix={arXiv},
      primaryClass={cs.AI},
      url={https://arxiv.org/abs/2503.19470}, 
}

@misc{song2025r1searcherincentivizingsearchcapability,
      title={R1-Searcher: Incentivizing the Search Capability in LLMs via Reinforcement Learning}, 
      author={Huatong Song and Jinhao Jiang and Yingqian Min and Jie Chen and Zhipeng Chen and Wayne Xin Zhao and Lei Fang and Ji-Rong Wen},
      year={2025},
      eprint={2503.05592},
      archivePrefix={arXiv},
      primaryClass={cs.AI},
      url={https://arxiv.org/abs/2503.05592}, 
}

@misc{qian2025toolrlrewardtoollearning,
      title={ToolRL: Reward is All Tool Learning Needs}, 
      author={Cheng Qian and Emre Can Acikgoz and Qi He and Hongru Wang and Xiusi Chen and Dilek Hakkani-Tür and Gokhan Tur and Heng Ji},
      year={2025},
      eprint={2504.13958},
      archivePrefix={arXiv},
      primaryClass={cs.LG},
      url={https://arxiv.org/abs/2504.13958}, 
}

@misc{li2025torlscalingtoolintegratedrl,
      title={ToRL: Scaling Tool-Integrated RL}, 
      author={Xuefeng Li and Haoyang Zou and Pengfei Liu},
      year={2025},
      eprint={2503.23383},
      archivePrefix={arXiv},
      primaryClass={cs.CL},
      url={https://arxiv.org/abs/2503.23383}, 
}

@inproceedings{su2024dragin,
  title={Dragin: Dynamic retrieval augmented generation based on the real-time information needs of large language models},
  author={Su, Weihang and Tang, Yichen and Ai, Qingyao and Wu, Zhijing and Liu, Yiqun},
  booktitle={Proceedings of the 62nd Annual Meeting of the Association for Computational Linguistics (Volume 1: Long Papers)},
  pages={12991--13013},
  year={2024}
}

@misc{zhang2026credibleplandrivenragmethod,
      title={Credible Plan-Driven RAG Method for Multi-Hop Question Answering}, 
      author={Ningning Zhang and Chi Zhang and Zhizhong Tan and Xingxing Yang and Weiping Deng and Wenyong Wang},
      year={2026},
      eprint={2504.16787},
      archivePrefix={arXiv},
      primaryClass={cs.CL},
      url={https://arxiv.org/abs/2504.16787}, 
}

@misc{dhole2025retrieveretrieveuncertaintydetection,
      title={To Retrieve or Not to Retrieve? Uncertainty Detection for Dynamic Retrieval Augmented Generation}, 
      author={Kaustubh D. Dhole},
      year={2025},
      eprint={2501.09292},
      archivePrefix={arXiv},
      primaryClass={cs.CL},
      url={https://arxiv.org/abs/2501.09292}, 
}

@inproceedings{jeong2024adaptive,
  title={Adaptive-rag: Learning to adapt retrieval-augmented large language models through question complexity},
  author={Jeong, Soyeong and Baek, Jinheon and Cho, Sukmin and Hwang, Sung Ju and Park, Jong C},
  booktitle={Proceedings of the 2024 Conference of the North American Chapter of the Association for Computational Linguistics: Human Language Technologies (Volume 1: Long Papers)},
  pages={7036--7050},
  year={2024}
}

@misc{zubkova2025sugarleveragingcontextualconfidence,
      title={SUGAR: Leveraging Contextual Confidence for Smarter Retrieval}, 
      author={Hanna Zubkova and Ji-Hoon Park and Seong-Whan Lee},
      year={2025},
      eprint={2501.04899},
      archivePrefix={arXiv},
      primaryClass={cs.CL},
      url={https://arxiv.org/abs/2501.04899}, 
}

@inproceedings{yao2025seakr,
  title={Seakr: Self-aware knowledge retrieval for adaptive retrieval augmented generation},
  author={Yao, Zijun and Qi, Weijian and Pan, Liangming and Cao, Shulin and Hu, Linmei and Weichuan, Liu and Hou, Lei and Li, Juanzi},
  booktitle={Proceedings of the 63rd Annual Meeting of the Association for Computational Linguistics (Volume 1: Long Papers)},
  pages={27022--27043},
  year={2025}
}

@inproceedings{huanshuo2025ctrla,
  title={Ctrla: Adaptive retrieval-augmented generation via inherent control},
  author={Huanshuo, Liu and Zhang, Hao and Guo, Zhijiang and Wang, Jing and Dong, Kuicai and Li, Xiangyang and Lee, Yi Quan and Zhang, Cong and Liu, Yong},
  booktitle={Findings of the Association for Computational Linguistics: ACL 2025},
  pages={12592--12618},
  year={2025}
}

@misc{sha2025semreinforcementlearningsearchefficient,
      title={SEM: Reinforcement Learning for Search-Efficient Large Language Models}, 
      author={Zeyang Sha and Shiwen Cui and Weiqiang Wang},
      year={2025},
      eprint={2505.07903},
      archivePrefix={arXiv},
      primaryClass={cs.CL},
      url={https://arxiv.org/abs/2505.07903}, 
}

@misc{wu2025searchwiselymitigatingsuboptimal,
      title={Search Wisely: Mitigating Sub-optimal Agentic Searches By Reducing Uncertainty}, 
      author={Peilin Wu and Mian Zhang and Xinlu Zhang and Xinya Du and Zhiyu Zoey Chen},
      year={2025},
      eprint={2505.17281},
      archivePrefix={arXiv},
      primaryClass={cs.CL},
      url={https://arxiv.org/abs/2505.17281}, 
}

@misc{song2025r1searcherincentivizingdynamicknowledge,
      title={R1-Searcher++: Incentivizing the Dynamic Knowledge Acquisition of LLMs via Reinforcement Learning}, 
      author={Huatong Song and Jinhao Jiang and Wenqing Tian and Zhipeng Chen and Yuhuan Wu and Jiahao Zhao and Yingqian Min and Wayne Xin Zhao and Lei Fang and Ji-Rong Wen},
      year={2025},
      eprint={2505.17005},
      archivePrefix={arXiv},
      primaryClass={cs.CL},
      url={https://arxiv.org/abs/2505.17005}, 
}

@misc{sun2026autosearchadaptivesearchdepth,
      title={AutoSearch: Adaptive Search Depth for Efficient Agentic RAG via Reinforcement Learning}, 
      author={Jingbo Sun and Wenyue Chong and Songjun Tu and Qichao Zhang and Yaocheng Zhang and Jiajun Chai and Xiaohan Wang and Wei Lin and Guojun Yin and Dongbin Zhao},
      year={2026},
      eprint={2604.17337},
      archivePrefix={arXiv},
      primaryClass={cs.AI},
      url={https://arxiv.org/abs/2604.17337}, 
}

@inproceedings{sun2025rearter,
  title={Rearter: Retrieval-augmented reasoning with trustworthy process rewarding},
  author={Sun, Zhongxiang and Wang, Qipeng and Yu, Weijie and Zang, Xiaoxue and Zheng, Kai and Xu, Jun and Zhang, Xiao and Song, Yang and Li, Han},
  booktitle={Proceedings of the 48th International ACM SIGIR Conference on Research and Development in Information Retrieval},
  pages={1251--1261},
  year={2025}
}

@misc{wu2026hipraghierarchicalprocessrewards,
      title={HiPRAG: Hierarchical Process Rewards for Efficient Agentic Retrieval Augmented Generation}, 
      author={Peilin Wu and Mian Zhang and Kun Wan and Wentian Zhao and Kaiyu He and Xinya Du and Zhiyu Chen},
      year={2026},
      eprint={2510.07794},
      archivePrefix={arXiv},
      primaryClass={cs.CL},
      url={https://arxiv.org/abs/2510.07794}, 
}

@misc{ye2025correctnessharmonizingprocessoutcome,
      title={Beyond Correctness: Harmonizing Process and Outcome Rewards through RL Training}, 
      author={Chenlu Ye and Zhou Yu and Ziji Zhang and Hao Chen and Narayanan Sadagopan and Jing Huang and Tong Zhang and Anurag Beniwal},
      year={2025},
      eprint={2509.03403},
      archivePrefix={arXiv},
      primaryClass={cs.LG},
      url={https://arxiv.org/abs/2509.03403}, 
}

@misc{yue2025promotingefficientreasoningverifiable,
      title={Promoting Efficient Reasoning with Verifiable Stepwise Reward}, 
      author={Chuhuai Yue and Chengqi Dong and Yinan Gao and Hang He and Jiajun Chai and Guojun Yin and Wei Lin},
      year={2025},
      eprint={2508.10293},
      archivePrefix={arXiv},
      primaryClass={cs.AI},
      url={https://arxiv.org/abs/2508.10293}, 
}

@article{guo2025deepseek,
  title={Deepseek-r1: Incentivizing reasoning capability in llms via reinforcement learning},
  author={Guo, Daya and Yang, Dejian and Zhang, Haowei and Song, Junxiao and Wang, Peiyi and Zhu, Qihao and Xu, Runxin and Zhang, Ruoyu and Ma, Shirong and Bi, Xiao and others},
  journal={arXiv preprint arXiv:2501.12948},
  year={2025}
}

@article{yang2025qwen3,
  title={Qwen3 technical report},
  author={Yang, An and Li, Anfeng and Yang, Baosong and Zhang, Beichen and Hui, Binyuan and Zheng, Bo and Yu, Bowen and Gao, Chang and Huang, Chengen and Lv, Chenxu and others},
  journal={arXiv preprint arXiv:2505.09388},
  year={2025}
}

@article{jaech2024openai,
  title={Openai o1 system card},
  author={Jaech, Aaron and Kalai, Adam and Lerer, Adam and Richardson, Adam and El-Kishky, Ahmed and Low, Aiden and Helyar, Alec and Madry, Aleksander and Beutel, Alex and Carney, Alex and others},
  journal={arXiv preprint arXiv:2412.16720},
  year={2024}
}

@article{ji2023survey,
  title={Survey of hallucination in natural language generation},
  author={Ji, Ziwei and Lee, Nayeon and Frieske, Rita and Yu, Tiezheng and Su, Dan and Xu, Yan and Ishii, Etsuko and Bang, Ye Jin and Madotto, Andrea and Fung, Pascale},
  journal={ACM computing surveys},
  volume={55},
  number={12},
  pages={1--38},
  year={2023},
  publisher={ACM New York, NY}
}

@inproceedings{kandpal2023large,
  title={Large language models struggle to learn long-tail knowledge},
  author={Kandpal, Nikhil and Deng, Haikang and Roberts, Adam and Wallace, Eric and Raffel, Colin},
  booktitle={International conference on machine learning},
  pages={15696--15707},
  year={2023},
  organization={PMLR}
}

@misc{liu2026bapoboundaryawarepolicyoptimization,
      title={BAPO: Boundary-Aware Policy Optimization for Reliable Agentic Search}, 
      author={Shiyu Liu and Yongjing Yin and Jianhao Yan and Yunbo Tang and Qinggang Zhang and Bei Li and Xin Chen and Jingang Wang and Xunliang Cai and Jinsong Su},
      year={2026},
      eprint={2601.11037},
      archivePrefix={arXiv},
      primaryClass={cs.AI},
      url={https://arxiv.org/abs/2601.11037}, 
}

@misc{karpukhin2020densepassageretrievalopendomain,
      title={Dense Passage Retrieval for Open-Domain Question Answering}, 
      author={Vladimir Karpukhin and Barlas Oğuz and Sewon Min and Patrick Lewis and Ledell Wu and Sergey Edunov and Danqi Chen and Wen-tau Yih},
      year={2020},
      eprint={2004.04906},
      archivePrefix={arXiv},
      primaryClass={cs.CL},
      url={https://arxiv.org/abs/2004.04906}, 
}

@misc{wang2024textembeddingsweaklysupervisedcontrastive,
      title={Text Embeddings by Weakly-Supervised Contrastive Pre-training}, 
      author={Liang Wang and Nan Yang and Xiaolong Huang and Binxing Jiao and Linjun Yang and Daxin Jiang and Rangan Majumder and Furu Wei},
      year={2024},
      eprint={2212.03533},
      archivePrefix={arXiv},
      primaryClass={cs.CL},
      url={https://arxiv.org/abs/2212.03533}, 
}

@article{kwiatkowski2019natural,
  title={Natural questions: a benchmark for question answering research},
  author={Kwiatkowski, Tom and Palomaki, Jennimaria and Redfield, Olivia and Collins, Michael and Parikh, Ankur and Alberti, Chris and Epstein, Danielle and Polosukhin, Illia and Devlin, Jacob and Lee, Kenton and others},
  journal={Transactions of the Association for Computational Linguistics},
  volume={7},
  pages={453--466},
  year={2019},
  publisher={MIT Press One Rogers Street, Cambridge, MA 02142-1209, USA journals-info~…}
}

@misc{joshi2017triviaqalargescaledistantly,
      title={TriviaQA: A Large Scale Distantly Supervised Challenge Dataset for Reading Comprehension}, 
      author={Mandar Joshi and Eunsol Choi and Daniel S. Weld and Luke Zettlemoyer},
      year={2017},
      eprint={1705.03551},
      archivePrefix={arXiv},
      primaryClass={cs.CL},
      url={https://arxiv.org/abs/1705.03551}, 
}

@misc{mallen2023trustlanguagemodelsinvestigating,
      title={When Not to Trust Language Models: Investigating Effectiveness of Parametric and Non-Parametric Memories}, 
      author={Alex Mallen and Akari Asai and Victor Zhong and Rajarshi Das and Daniel Khashabi and Hannaneh Hajishirzi},
      year={2023},
      eprint={2212.10511},
      archivePrefix={arXiv},
      primaryClass={cs.CL},
      url={https://arxiv.org/abs/2212.10511}, 
}

@misc{yang2018hotpotqadatasetdiverseexplainable,
      title={HotpotQA: A Dataset for Diverse, Explainable Multi-hop Question Answering}, 
      author={Zhilin Yang and Peng Qi and Saizheng Zhang and Yoshua Bengio and William W. Cohen and Ruslan Salakhutdinov and Christopher D. Manning},
      year={2018},
      eprint={1809.09600},
      archivePrefix={arXiv},
      primaryClass={cs.CL},
      url={https://arxiv.org/abs/1809.09600}, 
}

@misc{ho2020constructingmultihopqadataset,
      title={Constructing A Multi-hop QA Dataset for Comprehensive Evaluation of Reasoning Steps}, 
      author={Xanh Ho and Anh-Khoa Duong Nguyen and Saku Sugawara and Akiko Aizawa},
      year={2020},
      eprint={2011.01060},
      archivePrefix={arXiv},
      primaryClass={cs.CL},
      url={https://arxiv.org/abs/2011.01060}, 
}

@misc{trivedi2022musiquemultihopquestionssinglehop,
      title={MuSiQue: Multihop Questions via Single-hop Question Composition}, 
      author={Harsh Trivedi and Niranjan Balasubramanian and Tushar Khot and Ashish Sabharwal},
      year={2022},
      eprint={2108.00573},
      archivePrefix={arXiv},
      primaryClass={cs.CL},
      url={https://arxiv.org/abs/2108.00573}, 
}

@misc{press2023measuringnarrowingcompositionalitygap,
      title={Measuring and Narrowing the Compositionality Gap in Language Models}, 
      author={Ofir Press and Muru Zhang and Sewon Min and Ludwig Schmidt and Noah A. Smith and Mike Lewis},
      year={2023},
      eprint={2210.03350},
      archivePrefix={arXiv},
      primaryClass={cs.CL},
      url={https://arxiv.org/abs/2210.03350}, 
}

@misc{ahn2024largelanguagemodelsmathematical,
      title={Large Language Models for Mathematical Reasoning: Progresses and Challenges}, 
      author={Janice Ahn and Rishu Verma and Renze Lou and Di Liu and Rui Zhang and Wenpeng Yin},
      year={2024},
      eprint={2402.00157},
      archivePrefix={arXiv},
      primaryClass={cs.CL},
      url={https://arxiv.org/abs/2402.00157}, 
}

@misc{asai2023selfraglearningretrievegenerate,
      title={Self-RAG: Learning to Retrieve, Generate, and Critique through Self-Reflection}, 
      author={Akari Asai and Zeqiu Wu and Yizhong Wang and Avirup Sil and Hannaneh Hajishirzi},
      year={2023},
      eprint={2310.11511},
      archivePrefix={arXiv},
      primaryClass={cs.CL},
      url={https://arxiv.org/abs/2310.11511}, 
}

@misc{jiang2023activeretrievalaugmentedgeneration,
      title={Active Retrieval Augmented Generation}, 
      author={Zhengbao Jiang and Frank F. Xu and Luyu Gao and Zhiqing Sun and Qian Liu and Jane Dwivedi-Yu and Yiming Yang and Jamie Callan and Graham Neubig},
      year={2023},
      eprint={2305.06983},
      archivePrefix={arXiv},
      primaryClass={cs.CL},
      url={https://arxiv.org/abs/2305.06983}, 
}

@misc{zheng2025deepresearcherscalingdeepresearch,
      title={DeepResearcher: Scaling Deep Research via Reinforcement Learning in Real-world Environments}, 
      author={Yuxiang Zheng and Dayuan Fu and Xiangkun Hu and Xiaojie Cai and Lyumanshan Ye and Pengrui Lu and Pengfei Liu},
      year={2025},
      eprint={2504.03160},
      archivePrefix={arXiv},
      primaryClass={cs.AI},
      url={https://arxiv.org/abs/2504.03160}, 
}

@misc{li2025webthinkerempoweringlargereasoning,
      title={WebThinker: Empowering Large Reasoning Models with Deep Research Capability}, 
      author={Xiaoxi Li and Jiajie Jin and Guanting Dong and Hongjin Qian and Yongkang Wu and Ji-Rong Wen and Yutao Zhu and Zhicheng Dou},
      year={2025},
      eprint={2504.21776},
      archivePrefix={arXiv},
      primaryClass={cs.CL},
      url={https://arxiv.org/abs/2504.21776}, 
}

@misc{feng2025retoolreinforcementlearningstrategic,
      title={ReTool: Reinforcement Learning for Strategic Tool Use in LLMs}, 
      author={Jiazhan Feng and Shijue Huang and Xingwei Qu and Ge Zhang and Yujia Qin and Baoquan Zhong and Chengquan Jiang and Jinxin Chi and Wanjun Zhong},
      year={2025},
      eprint={2504.11536},
      archivePrefix={arXiv},
      primaryClass={cs.CL},
      url={https://arxiv.org/abs/2504.11536}, 
}

@misc{dao2025rezeroenhancingllmsearch,
      title={ReZero: Enhancing LLM search ability by trying one-more-time}, 
      author={Alan Dao and Thinh Le},
      year={2025},
      eprint={2504.11001},
      archivePrefix={arXiv},
      primaryClass={cs.CL},
      url={https://arxiv.org/abs/2504.11001}, 
}

@misc{jiang2025deepretrievalhackingrealsearch,
      title={DeepRetrieval: Hacking Real Search Engines and Retrievers with Large Language Models via Reinforcement Learning}, 
      author={Pengcheng Jiang and Jiacheng Lin and Lang Cao and Runchu Tian and SeongKu Kang and Zifeng Wang and Jimeng Sun and Jiawei Han},
      year={2025},
      eprint={2503.00223},
      archivePrefix={arXiv},
      primaryClass={cs.IR},
      url={https://arxiv.org/abs/2503.00223}, 
}

@misc{sun2026zerosearchincentivizesearchcapability,
      title={ZeroSearch: Incentivize the Search Capability of LLMs without Searching}, 
      author={Hao Sun and Zile Qiao and Jiayan Guo and Xuanbo Fan and Yingyan Hou and Yong Jiang and Pengjun Xie and Yan Zhang and Fei Huang and Jingren Zhou},
      year={2026},
      eprint={2505.04588},
      archivePrefix={arXiv},
      primaryClass={cs.CL},
      url={https://arxiv.org/abs/2505.04588}, 
}

@misc{fan2025ssrlselfsearchreinforcementlearning,
      title={SSRL: Self-Search Reinforcement Learning}, 
      author={Yuchen Fan and Kaiyan Zhang and Heng Zhou and Yuxin Zuo and Yanxu Chen and Yu Fu and Xinwei Long and Xuekai Zhu and Che Jiang and Yuchen Zhang and Li Kang and Gang Chen and Cheng Huang and Zhizhou He and Bingning Wang and Lei Bai and Ning Ding and Bowen Zhou},
      year={2025},
      eprint={2508.10874},
      archivePrefix={arXiv},
      primaryClass={cs.CL},
      url={https://arxiv.org/abs/2508.10874}, 
}

@misc{wu2025masksearchuniversalpretrainingframework,
      title={MaskSearch: A Universal Pre-Training Framework to Enhance Agentic Search Capability}, 
      author={Weiqi Wu and Xin Guan and Shen Huang and Yong Jiang and Pengjun Xie and Fei Huang and Jiuxin Cao and Hai Zhao and Jingren Zhou},
      year={2025},
      eprint={2505.20285},
      archivePrefix={arXiv},
      primaryClass={cs.CL},
      url={https://arxiv.org/abs/2505.20285}, 
}

@misc{zhang2025kbmdelineatingknowledgeboundary,
      title={KBM: Delineating Knowledge Boundary for Adaptive Retrieval in Large Language Models}, 
      author={Zhen Zhang and Xinyu Wang and Yong Jiang and Zile Qiao and Zhuo Chen and Guangyu Li and Feiteng Mu and Mengting Hu and Pengjun Xie and Fei Huang},
      year={2025},
      eprint={2411.06207},
      archivePrefix={arXiv},
      primaryClass={cs.CL},
      url={https://arxiv.org/abs/2411.06207}, 
}

@misc{huang2025reinforcedinternalexternalknowledgesynergistic,
      title={Reinforced Internal-External Knowledge Synergistic Reasoning for Efficient Adaptive Search Agent}, 
      author={Ziyang Huang and Xiaowei Yuan and Yiming Ju and Jun Zhao and Kang Liu},
      year={2025},
      eprint={2505.07596},
      archivePrefix={arXiv},
      primaryClass={cs.CL},
      url={https://arxiv.org/abs/2505.07596}, 
}

@misc{jiang2025s3dontneeddata,
      title={s3: You Don't Need That Much Data to Train a Search Agent via RL}, 
      author={Pengcheng Jiang and Xueqiang Xu and Jiacheng Lin and Jinfeng Xiao and Zifeng Wang and Jimeng Sun and Jiawei Han},
      year={2025},
      eprint={2505.14146},
      archivePrefix={arXiv},
      primaryClass={cs.AI},
      url={https://arxiv.org/abs/2505.14146}, 
}

@article{zhuang2025linearrag,
  title={Linearrag: Linear graph retrieval augmented generation on large-scale corpora},
  author={Zhuang, Luyao and Chen, Shengyuan and Xiao, Yilin and Zhou, Huachi and Zhang, Yujing and Chen, Hao and Zhang, Qinggang and Huang, Xiao},
  journal={arXiv preprint arXiv:2510.10114},
  year={2025}
}

@article{zhang2025survey,
  title={A survey of graph retrieval-augmented generation for customized large language models},
  author={Zhang, Qinggang and Chen, Shengyuan and Bei, Yuanchen and Yuan, Zheng and Zhou, Huachi and Hong, Zijin and Chen, Hao and Xiao, Yilin and Zhou, Chuang and Dong, Junnan and others},
  journal={arXiv preprint arXiv:2501.13958},
  year={2025}
}

@article{xiang2025use,
  title={When to use graphs in rag: A comprehensive analysis for graph retrieval-augmented generation},
  author={Xiang, Zhishang and Wu, Chuanjie and Zhang, Qinggang and Chen, Shengyuan and Hong, Zijin and Huang, Xiao and Su, Jinsong},
  journal={arXiv preprint arXiv:2506.05690},
  year={2025}
}

@article{xiao2026reliable,
  title={Reliable reasoning path: Distilling effective guidance for llm reasoning with knowledge graphs},
  author={Xiao, Yilin and Zhou, Chuang and Zhang, Qinggang and Li, Bo and Li, Qing and Huang, Xiao},
  journal={IEEE Transactions on Knowledge and Data Engineering},
  year={2026},
  publisher={IEEE}
}

@inproceedings{zhang2025faithfulrag,
  title={Faithfulrag: Fact-level conflict modeling for context-faithful retrieval-augmented generation},
  author={Zhang, Qinggang and Xiang, Zhishang and Xiao, Yilin and Wang, Le and Li, Junhui and Wang, Xinrun and Su, Jinsong},
  booktitle={Proceedings of the 63rd Annual Meeting of the Association for Computational Linguistics (Volume 1: Long Papers)},
  pages={21863--21882},
  year={2025}
}

@article{gao2025probing,
  title={Probing Latent Knowledge Conflict for Faithful Retrieval-Augmented Generation},
  author={Gao, Linfeng and Bi, Baolong and Yuan, Zheng and Wang, Le and Chen, Zerui and Wei, Zhimin and Liu, Shenghua and Zhang, Qinggang and Su, Jinsong},
  journal={arXiv preprint arXiv:2510.12460},
  year={2025}
}

@article{xiang2026systematic,
  title={A Systematic Survey of Self-Evolving Agents: From Model-Centric to Environment-Driven Co-Evolution},
  author={Xiang, Zhishang and Yang, Chengyi and Chen, Zerui and Wei, Zhimin and Tang, Yunbo and Teng, Zongpei and Peng, Zexi and Li, Zongxia and Huang, Chengsong and He, Yicheng and others},
  year={2026},
  publisher={TechRxiv}
}

@article{yang2026graph,
  title={Graph-based Agent Memory: Taxonomy, Techniques, and Applications},
  author={Yang, Chang and Zhou, Chuang and Xiao, Yilin and Dong, Su and Zhuang, Luyao and Zhang, Yujing and Wang, Zhu and Hong, Zijin and Yuan, Zheng and Xiang, Zhishang and others},
  journal={arXiv preprint arXiv:2602.05665},
  year={2026}
}

@inproceedings{chen2026you,
  title={You don’t need pre-built graphs for rag: Retrieval augmented generation with adaptive reasoning structures},
  author={Chen, Shengyuan and Zhou, Chuang and Yuan, Zheng and Zhang, Qinggang and Cui, Zeyang and Chen, Hao and Xiao, Yilin and Cao, Jiannong and Huang, Xiao},
  booktitle={Proceedings of the AAAI Conference on Artificial Intelligence},
  volume={40},
  number={36},
  pages={30270--30278},
  year={2026}
}
